\documentclass[journal,onecolumn]{IEEEtran}

\usepackage{url,hyperref,lineno}

\usepackage{amsmath,graphicx}

\author{Nicolas~Skatchkovsky, Hyeryung Jang, and Osvaldo~Simeone}

\title{Bayesian Continual Learning via Spiking Neural Networks}

\usepackage{booktabs} 
\usepackage{amsfonts}
\usepackage{amssymb}
\usepackage{amsthm}
\usepackage{pifont}
\usepackage{color}
\usepackage{array}
\usepackage{cite}
\usepackage{enumerate}
\usepackage{graphicx}
\usepackage{epstopdf}
\usepackage{epsfig}
\usepackage{footmisc}
\usepackage{amsbsy}
\usepackage{bm}
\usepackage{cases}
\usepackage{multirow}
\usepackage{algorithmic}
\usepackage{comment}
\usepackage{times}
\usepackage{paralist}
\usepackage[ruled]{algorithm2e}
\usepackage{setspace}
\usepackage{mathtools}

\newcommand{\bmx}{{\bm x}}
\newcommand{\bmy}{{\bm y}}

\newcommand{\bms}{{\bm s}}

\newcommand{\bma}{{\bm a}}

\newcommand{\bmw}{{\bm w}}

\newcommand{\set}[1]{\ensuremath{\mathcal #1}}

\usepackage{setspace}
\usepackage{lscape}

\begin{document}

\maketitle

\begin{abstract}

Among the main features of biological intelligence are energy efficiency, capacity for continual adaptation, and risk management via uncertainty quantification. Neuromorphic engineering has been thus far mostly driven by the goal of implementing energy-efficient machines that take inspiration from the time-based computing paradigm of biological brains. In this paper, we take steps towards the design of neuromorphic systems that are capable of adaptation to changing learning tasks, while producing well-calibrated uncertainty quantification estimates. To this end, we derive online learning rules for  spiking neural networks (SNNs)  within a Bayesian continual learning framework. In it, each synaptic weight is represented by parameters that quantify the current epistemic uncertainty resulting from prior knowledge and observed data. The proposed online rules update the distribution parameters in a streaming fashion as data are observed. We instantiate the proposed approach for both real-valued and binary synaptic weights. Experimental results using Intel's Lava platform show the merits of Bayesian over frequentist learning in terms of capacity for adaptation and uncertainty quantification.

\end{abstract}




\section{Introduction}

Recent advances in machine learning and artificial intelligence systems have been largely driven by a pursuit of accuracy via resource-intensive pattern recognition algorithms run in a \emph{train-and-then-deploy} fashion. In stark contrast, neuroscience paints a picture of intelligence that revolves around \emph{continual adaptation}, \emph{uncertainty quantification}, and resource budgeting (allostasis) for the parsimonious processing of \emph{event-driven} information  \cite{friston10free, doya07bayesian, athousbrain, sevenhalf}. Taking inspiration from neuroscience, over the last decade, neuromorphic engineering has pursued the goal of implementing energy-efficient machines that process information \emph{with} time via sparse inter-neuron binary signals -- or \emph{spikes} \cite{davies21loihiresults}. The main aim of this paper is to introduce algorithmic solutions to endow neuromorphic models, namely spiking neural networks (SNNs), with the capacity for adaptation to changing learning tasks, while ensuring the reliable quantification of uncertainty of the model's decisions. 

\subsection{Managing Uncertainty via Bayesian Learning}

Training algorithms for SNNs have been overwhelmingly derived by following the \emph{frequentist} approach which consists in minimizing the training loss with respect to the model parameter vector \cite{bellec20eprop, kaiser2020decolle, zenke2018superspike, shrestha18slayer}. This is partly motivated by the dominance of frequentist learning, and associated software tools, in the literature on deep learning for conventional artificial neural networks (ANNs). Frequentist learning is well justified when enough data are available to make the training loss a good empirical approximation of the underlying population loss \cite{clayton21bernoulli}. When this condition is not satisfied, while the model's average accuracy may be satisfactory on test data, the decisions made by the trained model can be badly calibrated, often resulting in overconfident predictions \cite{nguyen15overconfidence, guo2017hierarchical}. The problem is particularly significant for decisions made on test data that differ significantly from the data observed during training -- a common occurrence for applications such as self-driving vehicles. Furthermore, the inability of frequentist learning to account for uncertainty limits its capacity to adapt to new tasks while retaining the capacity to operate on previous tasks \cite{ebrahimi20uncertaintyguided}.

The main cause of the poor calibration of frequentist learning is the selection of a single parameter vector, which disregards any uncertainty on the best model to use for a certain task due to the availability of limited data. A more principled approach that has the potential to properly account for such \emph{epistemic uncertainty}, i.e., for uncertainty related to the availability of limited data, is given by \emph{Bayesian learning} \cite{jaynes03probability} and by its generalized form known as \emph{information risk minimization} (see, e.g., \cite{zhang2006itbound, guedj2019primer, knoblauch2019generalized, jose20free, simeone22ml4eng}). Bayesian learning maintains a \emph{distribution} over the model parameter vector that represents the partial information available to the learner. This way, Bayesian models can provide well-calibrated decisions, which quantify accurately the associated degree of uncertainty and can be used to detect out-of-distribution inputs \cite{daxberger22ood}. In the self-driving example provided earlier, the vehicle may hand back control to the driver when the certainty of its decision is below a certain threshold.

Bayesian reasoning is at the core of the \textit{Bayesian brain} hypothesis in neuroscience, according to which biological brains constantly update an internal model of the world in an attempt to minimize their information-theoretic surprise. This hypothesis is formalized by the free energy principle, which  measures surprise in terms of a variational free energy \cite{friston12bayesian}. In this context, synaptic plasticity has been hypothesized to be well-modelled as Bayesian learning, which keeps track of the distributions of synaptic weights over time \cite{aitchison21synapticbayesian}.

In the present paper, we propose (generalized) Bayesian learning rules for SNNs with binary and real-valued synaptic weights that can adapt over time to changing learning tasks.

\subsection{Related Work}

\label{sec:previous}

Bayesian learning, and its application to deep ANNs, typically labelled as \textit{Bayesian deep learning}, is receiving increasing attention in the literature. We refer to the following work for a recent overview \cite{wang21bayesian}. natural gradient descent rule known as the \textit{Bayesian learning rule} was introduced in \cite{khan2017conjugate}, then applied in \cite{meng2020bayesbinn} to train binary ANNs, and to a variety of other scenarios in \cite{khan21bayesian}. Reference \cite{khan21bayesian} demonstrates that the Bayesian learning rule recovers many state-of-the-art machine learning algorithms in a principled fashion. We also point to the reference \cite{kreutzer2020natural} that explores the use of natural gradient descent for frequentist learning in spiking neurons. 

As mentioned, the choice of a Bayesian learning framework is in line with the importance of the Bayesian brain hypothesis in computational neurosciences \cite{friston12bayesian}. The recent reference \cite{aitchison21synapticbayesian} explores a Bayesian paradigm to model biological synapses as an explanation of the capacity of the brain to perform learning in the presence of noisy observations. A Bayesian approach to neural plasticity was previously proposed for synaptic sampling, by modeling synaptic plasticity as sampling from a posterior distribution \cite{kappel15:synaptic}. Apart from the conference version \cite{jang21bisnn} of the present work, this paper is the first to explore the definition of Bayesian learning and Bayesian continual learning rules for general SNNs adopting the standard spike response model (SRM, see, e.g., \cite{gerstner2002spiking}).

Continual learning is a key area of machine learning research, which is partly motivated by the goal of understanding how biological brains maintain previously acquired skills while adding new capabilities. Unlike traditional machine learning, whereby one performs training based on a single data source, in continual learning, several datasets, corresponding to different tasks, are sequentially presented to the learner.  A challenge in continual learning is the ability of the learning algorithm to perform competitively on previous tasks after training on the subsequently observed datasets. In this context, \textit{catastrophic forgetting} indicates the situation in which performance drops sharply on previously encountered tasks after learning new ones. Many continual learning techniques follow the principle of preserving synaptic connections that are deemed important to perform well on previously learned tasks via a regularization of the learning objective \cite{kirkpatrick17ewc, zenke17continual}. Bayesian approaches have also been proposed for this purpose, whereby priors are selected as the posterior evaluated on the previous task to prevent the new posterior distribution from deviating too much from learned states. Biological mechanisms are explicitly leveraged in works such as  \cite{laborieux21metaplasticity} and \cite{soures21tacos}, which combine a variety of neural mechanisms to obtain state-of-the-art performance for SNNs on standard continual learning benchmarks. Reference \cite{putra22continual} also proposes a continual learning algorithm for SNNs in an unsupervised scenario by assuming limited precision for the weights. In the present paper, we demonstrate how Bayesian learning allows obtaining similar biologically inspired features by following a principled objective grounded in information risk minimization. 

Traditionally, training of SNNs has relied on biologically realistic Hebbian rules, among which spike-timing dependent plasticity (STDP) is the most popular. STDP modulates the synaptic weight between two neurons based on the firing times of both neurons. A long-term potentiation (i.e., an increase in the weight) of the synapse occurs when the pre-synaptic neuron spikes right before the post-synaptic neuron, while long-term depression (i.e., a decrease in the weight) of a synapse happens when the pre-synaptic neuron spikes after the post-synaptic neuron. STDP implements a form of unsupervised learning, and can be leveraged to perform tasks such as clustering, while also supporting continual learning \cite{vaila19deepsnn}.

Supervised learning based on the minimization of the training loss is challenging in SNNs due to the activation function of spiking neurons, the derivative of which is always zero, except at the spike time, where it is not differentiable. Modern training algorithms \cite{kaiser2020decolle, bellec20eprop, zenke2018superspike} overcome this difficulty through the use of \textit{surrogate gradients}, i.e., by replacing the true derivative with that of a well-defined differentiable function \cite{neftci2019surrogate}. An alternative approach, reviewed in \cite{jang19:spm}, is to view the SNN as a probabilistic model whose likelihood can be directly differentiated. Further extensions of the probabilistic modelling approach and associated training rules are presented in \cite{jang22multisample, jang20:vowel}.

An application of Bayesian principles to SNNs has first been proposed in the conference version of this paper \cite{jang21bisnn}. Reference  \cite{jang21bisnn} focuses on SNNs with binary synaptic weights and offline learning, presenting limited experimental results. In contrast, the current paper provides all the necessary background, including frequentist learning; it covers frequentist and Bayesian continual learning; and it provides extensive experimental results on a variety of tasks.

\subsection{Main Contributions}

In this work, we derive online learning rules for  SNNs  within a Bayesian continual learning framework. In it, each synaptic weight is represented by parameters that quantify the current epistemic uncertainty associated with prior knowledge and data observed thus far. Bayesian methods are key to handling uncertainty over time, providing the model knowledge of what is to be retained, and what can be forgotten \cite{ebrahimi20uncertaintyguided}. The main contributions are as follows. \newline
\textit{i)} We introduce general frameworks for the definition of single-task and continual Bayesian learning problems for SNNs that are based on information risk minimization and variational inference. Following the desiderata formulated in \cite{farquhar19robust}, we focus on the standard formulation of continual learning in which there exist clear demarcations between subsequent tasks, but the learner is unaware of the identity of the current task. For example, in the typical example of an autonomous vehicle navigating in several environments, the vehicle may be aware that it is encountering a new terrain, while being a priori unaware of the type of new terrain. Furthermore, the model is not modified between tasks, and tasks may be encountered more than once; \newline
\textit{ii)} We instantiate the general Bayesian learning frameworks for SNNs with real-valued synapses. To this end, we adopt a Gaussian variational distribution for the synaptic weights, and demonstrate learning rules that can adapt the parameters of the weight distributions online. This choice of variational posterior has been previously explored for ANNs, and can yield state-of-the-art performance on real-life datasets \cite{osawa2019practical}; \newline
\textit{iii)} We then introduce Bayesian single-task and continual learning rules for SNNs with binary weights, with the main goal of supporting more efficient hardware implementations \cite{courbariaux2016binarized, rastegari2016xnor}, including platforms based on beyond-CMOS memristors \cite{adnan}; \newline
\textit{iv)} Through experiments on both synthetic and real neuromorphic datasets, we demonstrate the advantage of the Bayesian learning paradigm in terms of accuracy and calibration for both single-task and continual learning. As neuromorphic algorithms are designed to be run on dedicated hardware, we run the experiments using Intel's Lava software emulator platform \cite{intel22lava}, accounting for the limited precision of synaptic weights in hardware.




\section{Methods}
We first introduce the adopted SNN model, namely the standard spike response model (SRM), before giving a short overview of frequentist, Bayesian, continual, and biologically inspired learning. We then detail learning rules for offline and continual frequentist learning, and derive associated online Bayesian learning rules.

\subsection{SNN Model}
\begin{figure}[t!]
\centering
\includegraphics[width=0.5\columnwidth]{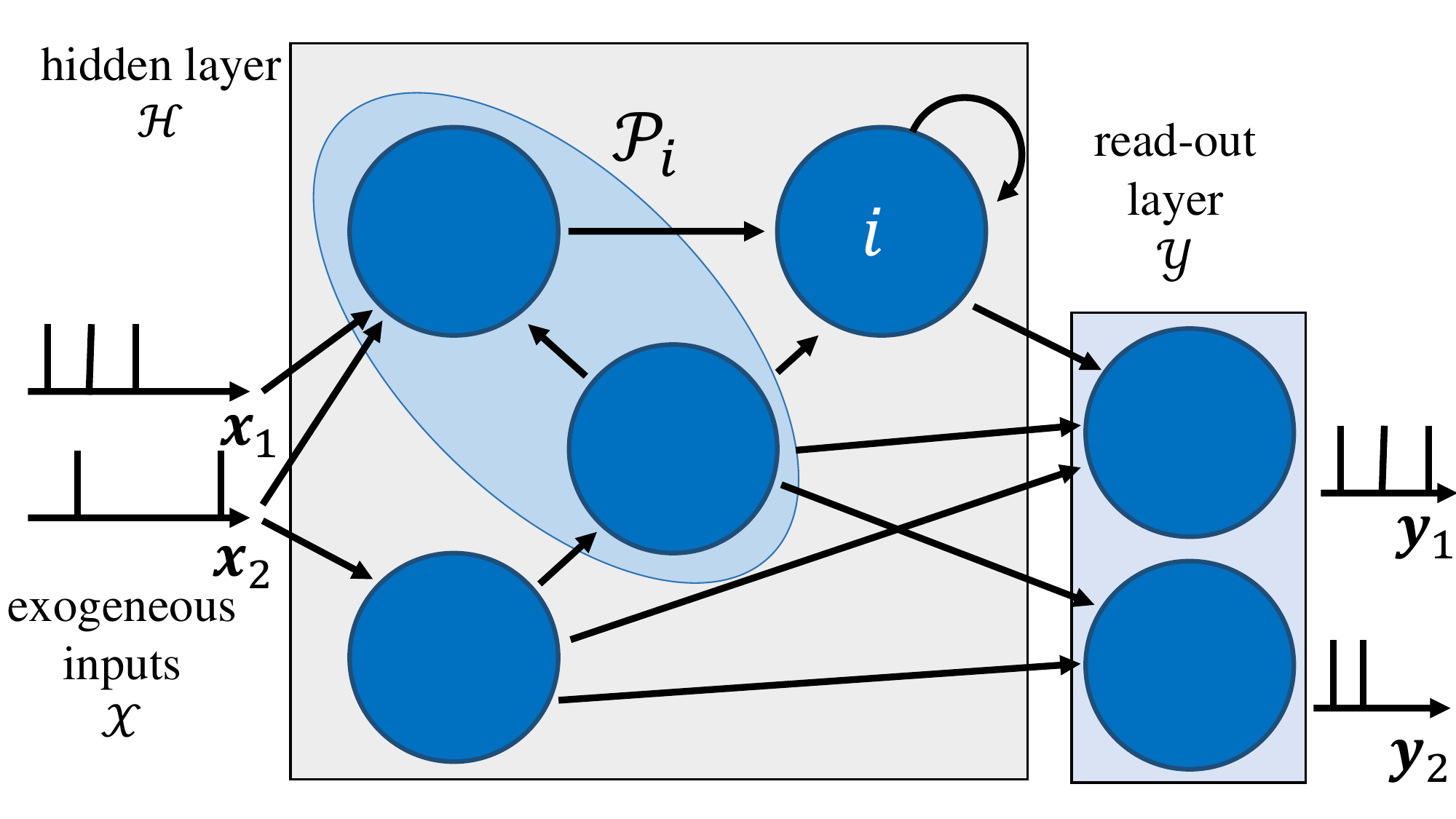}
\caption{Illustration of the internal architecture of an SNN. The behavior of neurons in the read-out layer is guided by the training data, while that of neurons in the hidden layer is adjusted to fit the data. The blue shaded area represents the set of pre-synaptic neurons $\set{P}_i$ to neuron $i$.}
\label{fig:snn}
\end{figure}

\subsubsection{Spike Response Model}
\label{sec:srm}
The architecture of an SNN is defined by a network of spiking neurons connected over an arbitrary graph, which possibly includes (directed) cycles. As illustrated in Fig.~\ref{fig:snn}, the directed graph $\mathcal{G}=(\set{N}, \set{E})$ is described by a set $\set{N}$ of nodes, representing the neurons, and by a set $\set{E}$ of directed edges $i \rightarrow j$ with $i \ne j \in \set{N}$, representing synaptic connections.

Focusing on a discrete-time implementation, each spiking neuron $i \in \set{N}$ produces a binary value $s_{i,t} \in \{0,1\}$ at discrete time $t=1,2,\ldots$, with ``$1$'' denoting the firing of a spike. 
We collect in an $|\set{N}| \times 1$ vector $\bms_t = (s_{i,t}: i \in \set{N})$ the spikes emitted by all neurons $\set{N}$ at time $t$, and denote by $\bms^{t} = (\bms_1, \ldots, \bms_t)$ the spike sequences of all neurons up to time $t$. Without loss of generality, we consider time-sequences of length $T$, and write  $\bms := \bms^T$.
Each neuron $i$ receives input spike signals $\{s_{j,t}\}_{j \in \set{P}_i} = \bms_{\set{P}_i, t}$ at time $t$ from the set $\set{P}_i = \{j \in \set{N}: (j \rightarrow i) \in \set{E}\}$ of parent, or pre-synaptic, neurons, which are connected to neuron $i$ via directed links in the graph $\mathcal{G}$. With some abuse of notations, this set is taken to include also exogeneous input signals. 

Each neuron $i$ maintains a scalar analog state variable $u_{i,t}$, known as the {\em membrane potential}. Mathematically, neuron $i$ outputs a binary signal $s_{i,t}$, or spike, at time $t$ when the membrane potential $u_{i,t}$ is above a threshold $\vartheta$, i.e.,

\begin{align} \label{eq:ind-threshold}
    s_{i,t} = \Theta(u_{i,t} - \vartheta),
\end{align}

with $\Theta(\cdot)$ being the Heaviside step function and $\vartheta$ being the fixed firing threshold. 
Following the standard discrete-time SRM \cite{gerstner2002spiking}, the membrane potential $u_{i,t}$ is obtained by summing filtered contributions from pre-synaptic neurons in set $\set{P}_i$ and from the neuron's own output. In particular, the membrane potential evolves as

\begin{align} \label{eq:ind-membrane}
    u_{i,t} = \sum_{j \in \set{P}_i} w_{ij} \big( {\alpha_t} \ast s_{j,t} \big) - \beta_t \ast s_{i,t},
\end{align}
where $w_{ij}$ is a learnable synaptic weight from pre-synaptic neuron $j \in \set{P}_i$ to post-synaptic neuron $i$; and we collect in vector $\bmw = \{\bmw_i\}_{i \in \set{N}}$ the model parameters, with $\bmw_i := \{w_{ij}\}_{j \in \set{P}_i}$ being the synaptic weights for each neuron $i$. We have denoted as $\alpha_t$ and $\beta_t$ the spike responses of synapses and somas, respectively; while $\ast$ denotes the convolution operator $f_t \ast g_t = \sum_{\delta > 0} f_{\delta} g_{t-\delta}$. When implemented with autoregressive filters, the SRM is equivalent to leaky integrate-and-fire (LIF) neuron model \cite{gerstner2002spiking, kaiser2020decolle}. The techniques developed in this work can be directly generalized to other, more complex, neuron models, such as resonate-and-fire \cite{izhikevich01raf}, but we leave an investigation of this point to future work.

\subsubsection{Real-Valued and Binary-Valued Synapses}
In this paper, we will consider two implementations of the SRM introduced in the previous subsection. In the first, the synaptic weights in vector $\bmw$ are real-valued, i.e., $w_{ij} \in \mathbb{R}$, with possibly limited resolution, as dictated by deployment on neuromorphic hardware (see Sec.~\ref{sec:exp}). In contrast, in the second implementation, the weights are binary, i.e., $w_{ij} \in \{+1, -1\}$. The advantages of models with binary-valued synapses, which we call binary SNNs, include a reduced complexity for the computation of the membrane potential $u_{i,t}$ in \eqref{eq:ind-membrane}. 
Furthermore, binary SNNs are particularly well suited for implementations on chips with nanoscale components that provide discrete conductance levels for the synapses \cite{adnan}. In this regard, we note that the methods described in this paper can be generalized to models with weights having any discrete number of values. 

\subsection{Frequentist versus Bayesian Learning}
\begin{figure}[t!]
\centering
\includegraphics[width=\columnwidth]{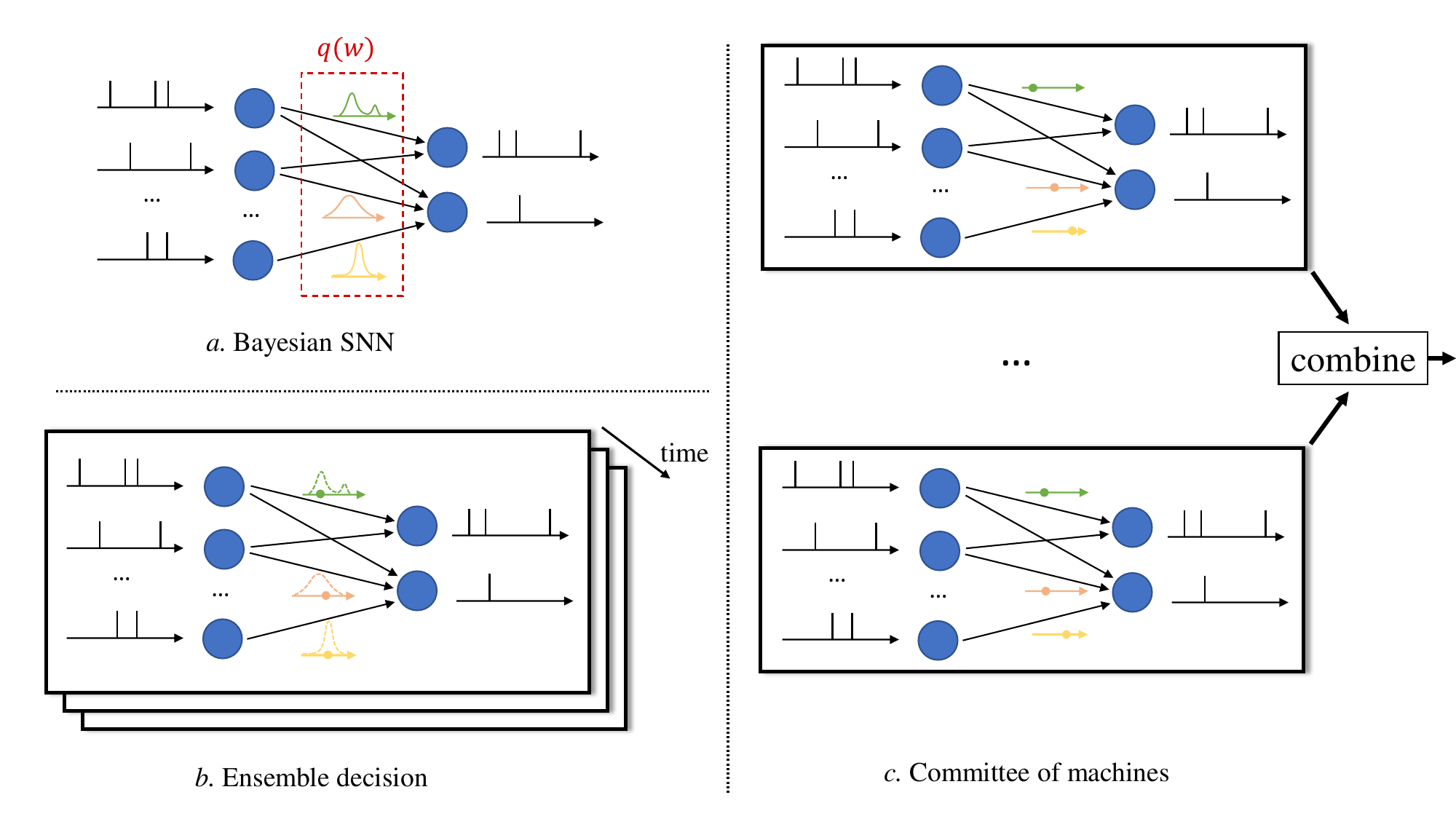} 
\caption{Illustration of Bayesian learning in an SNN. \textit{(a)} In a Bayesian SNN, the synaptic weights $\bmw$ are assigned a joint distribution $q(\bmw)$, often simplified as a product distribution across weights. \textit{(b)} An ensemble decision is obtained by sampling several times from the distribution $q(\bmw)$, and by averaging the predictions of the sampled models. Sampling is done independently for each new input. \textit{(c)} In a committee machine, the weights for several models are drawn only once from $q(\bmw)$, and the same models are run in parallel for any new input.}
\label{fig:overview}
\end{figure}

\label{sec:predictors}
With traditional frequentist learning, the vector of synaptic weights $\bmw$ is optimized by minimizing a training loss. The training loss is adopted as a proxy for the population loss, i.e., for the loss averaged over the true, unknown, distribution of the data. Therefore, frequentist learning disregards the inherent uncertainty caused by the availability of limited training data, which causes the training loss to be a potentially inaccurate estimate of the population loss. As a result, frequentist learning is known to potentially yield poorly calibrated, and overconfident decisions for ANNs \cite{nguyen15overconfidence}.

In contrast, as seen in Fig.~\ref{fig:overview}a, Bayesian learning optimizes over a distribution $q(\bmw)$ in the space of the synaptic weight vector $\bmw$. The distribution $q(\bmw)$ captures the \textit{epistemic} uncertainty induced by the lack of knowledge of the true distribution of the data. This is done by assigning similar values of $q(\bmw)$ to model parameters that fit equally well the data, while also being consistent with prior knowledge. As a consequence, Bayesian learning is known to produce better calibrated decisions, i.e., decisions whose associated confidence better reflects the actual accuracy of the decision \cite{guo2017hierarchical}. Furthermore, models trained via Bayesian learning can better detect out-of-distribution data, i.e., data that is not covered by the distribution of the training set \cite{kristiadi20bayesian, daxberger22ood}.

Once distribution $q(\bmw)$ is optimized via Bayesian learning, at inference time a decision on any new test input is made by averaging the decisions of multiple models, with each being drawn from the distribution $q(\bmw)$. 
The average over multiple models can be realized in one of two ways. \newline
\textit{i)} \textit{Ensemble predictor:} Given a test input, as seen in Fig.~\ref{fig:overview}b, one draws a new synaptic weight vector several times from the distribution $q(\bmw)$, and an \textit{ensemble} decision is obtained by averaging the decisions produced by running the SNN with each sampled weight vector; \newline
\textit{ii)} \textit{Committee machine:} Alternatively, one can sample a number of realizations from the distribution $q(\bmw)$ that are kept fixed and reused for all test inputs. This solution foregoes the sampling step at inference time as illustrated in Fig.~\ref{fig:overview}c. However, the approach generally requires a larger memory to store all samples $\bmw$ to be used for inference, while the ensemble predictor can make decisions using different weight vectors $\bmw \sim q(\bmw)$ sequentially over time. 

\begin{figure}[t!]
\centering
\includegraphics[width=\columnwidth]{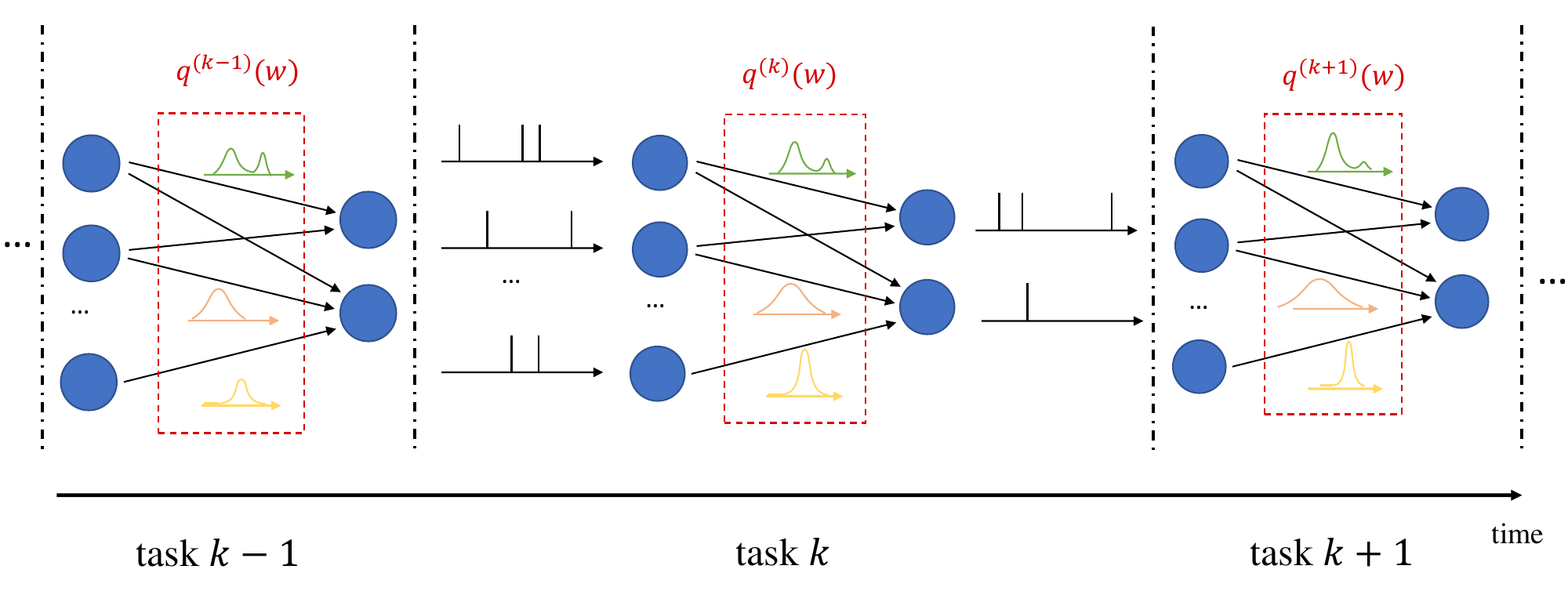}
\caption{Illustration of Bayesian continual learning: the system is successively presented with similar, but different, tasks. Bayesian learning allows the model to retain information about previously learned information.}
\label{fig:continual}
\end{figure}

\subsection{Offline versus Continual Learning}
Offline learning denotes the typical situation where the system is presented with a single training dataset $\set{D}$, which is used to measure a training loss. In offline learning, optimization of the training loss is carried out once and for all, resulting in a synaptic weight vector $\bmw$ or in a distribution $q(\bmw)$ for frequentist or Bayesian learning, respectively. Offline learning is hence, by construction, unable to adapt to changing conditions, and it is deemed to be a poor representation of how intelligence works in biological organisms \cite{kudithipudi22lifelong}.

In continual learning, the system is sequentially presented datasets $\mathcal{D}^{(1)}, \mathcal{D}^{(2)}, \dots$ corresponding to distinct, but related, learning tasks, where each task is selected, possibly with replacement, from a pool of tasks, and its identity is unknown to the system. For each task $k$, the system is given a training set $\mathcal{D}^{(k)}$, and its goal is to learn to make predictions that generalize well on the new task, while causing minimal loss of accuracy on previous tasks $1,\dots, k-1$. In frequentist continual learning, the model parameter vector $\bmw$ is updated as data from successive tasks is collected. Conversely, in Bayesian continual learning, the distribution $q(\bmw)$ is updated over time as illustrated in Fig.~\ref{fig:continual}. The updates should be sufficient to address the needs of the new task, while not disrupting performance on previous tasks, operating on a \textit{stability-plasticity trade-off}. 

\subsection{Biological Principles of Learning}
\label{sec:biological-learning}
Many existing works on continual learning draw their inspiration from the mechanisms underlying the capability of biological brains to carry out life-long learning \cite{soures21tacos, kudithipudi22lifelong}.  Learning is believed to be achieved in biological systems by modulating the strength of synaptic links. In this process, a variety of mechanisms are at work to establish short-to intermediate-term and long-term memory for the acquisition of new information over time \cite{kandel14memory}. These mechanisms operate at different time and spatial scales.

One of the best understood mechanisms, \textit{long-term potentiation}, contributes to the management of long-term memory through the consolidation of synaptic connections \cite{malenka04ltp, morris03memory}. Once established, these are rendered resistant to disruption by changing their capacity to change via \textit{metaplasticity} \cite{finnie12metaplasticity, abraham91metaplasticity}. As a related mechanism, return to a base state is ensured after exposition to small, noisy changes by \textit{heterosynaptic plasticity}, which plays a key role in ensuring the stability of neural systems \cite{chistiakova14heteroplasticity}. \textit{Neuromodulation} operates at the scale of neural populations to respond to particular events registered by the brain \cite{marder12neuromodulation}. Finally, \textit{episodic replay} plays a key role in the maintenance of long-term memory, by allowing biological brains to re-activate signals seen during previous active periods when inactive (i.e., sleeping) \cite{kudithipudi22lifelong}.





\subsection{Frequentist Offline Learning} 
\label{sec:freq-offline}

We now review frequentist offline training algorithms for SNNs, under the SRM model described in Sec.~\ref{sec:srm}. This will provide the necessary background for Bayesian learning and its continual version, described in Sec.~\ref{sec:bayes_snn} and Sec.~\ref{sec:bayes_cont_snn}, respectively.

\subsubsection{Empirical Risk Minimization}
\label{sec:erm}
To start, as illustrated in Fig.~\ref{fig:snn}, we divide the set $\set{N}$ of neurons of the SNN into two subsets $\set{Y}$ and $\set{H}$ with $\set{N} = \set{Y} \cup \set{H}$: a set of read-out, or output, neurons $\set{Y}$ and a set of hidden neurons $\set{H}$. The set of exogeneous inputs is defined as $\set{X}$. We focus on supervised learning, in which a dataset $\set{D}$ is given by $|\set{D}|$ pairs $(\bm{x}, \bm{y})$ of signals generated from an unknown distribution $p(\bm{x}, \bm{y})$, with $\bm{x}$ being exogeneous input signals, one for each element of the set $\set{X}$, and $\bm{y}$ the corresponding desired output signals. Both $\bmx$ and $\bmy$ are vector sequences of length $T$, with $\bmx$ comprising $|\set{X}|$ signals, and $\bmy$ including $|\set{Y}|$ signals. Each output samples $y_{m,t}$ in $\bmy$ dictates the desired behavior of the $m$th neuron in the read-out set $\set{Y}$. The sequences in $\bmx$ and $\bmy$ can generally take arbitrary real values (see Sec.~\ref{sec:exp} for specific examples).  

In frequentist learning, the goal is to minimize the training loss over the parameter vector $\bmw$ using the training dataset $\mathcal{D} = \{(\bm{x}, \bm{y})\}$. To elaborate, we define the loss $\set{L}_{\bm{x}, \bm{y}}(\bmw)$ measured with respect to a data $(\bm{x}, \bm{y}) \in \mathcal{D}$ as the error between the reference signals $\bm{y}$ and the output spiking signals produced by the SNN with parameters $\bmw$, given the input $\bm{x}$. 
Accordingly, the loss is written as a sum over time instants $t=1,\ldots,T$ and over the $|\set{Y}|$ read-out neurons as 

\begin{align} \label{eq:loss-decomp}
    \set{L}_{\bm{x}, \bm{y}}(\bmw) = \sum_{t=1}^T \set{L}_{\bm{x}^t, \bm{y}_t}(\bmw) = \sum_{t=1}^T \sum_{m \in \set{Y}} L\big(y_{m,t}, f_{m}(\bmw, \bmx^t)\big),
\end{align}
where function $ L\big(y_{m,t}, f_{m}(\bmw, \bmx^t)\big)$ is a local loss measure comparing the target output $y_{m,t}$ of neuron $m$ at time $t$ and the actual output $f_{m}(\bmw, \bmx^t)$ of the same neuron. The notations $f_{m}(\bmw, \bmx^t)$ and $\set{L}_{\bm{x}^t, \bm{y}_t}(\bmw)$ are used as a reminder that the output of the SNN and the corresponding loss at time $t$ generally depend on the input $\bmx^t$ up to time $t$, and on the target output $\bmy_t$ at time $t$. Specifically, the notation $f_{m}(\bmw, \bmx^t)$ makes it clear that the output of neuron $m \in \set{Y}$ is produced with the model parameters $\bmw$ from exogeneous input $\bm{x}^t$, consisting of all input samples up to time $t$, using the SRM \eqref{eq:ind-threshold}-\eqref{eq:ind-membrane}.

The training loss $\set{L}_{\set{D}}(\bmw)$ is an empirical estimate of the population loss based on the data samples in the training dataset $\mathcal{D}$, and is given as

\begin{align} \label{eq:loss}
     \mathcal{L}_{\set{D}}(\bmw) = \frac{1}{|\mathcal{D}|} \sum_{(\bm{x}, \bm{y}) \in \mathcal{D}} \mathcal{L}_{\bm{x}, \bm{y}}(\bmw).
\end{align}

Frequentist learning addresses the empirical risk minimization (ERM) problem

\begin{align} \label{eq:erm}
     \min_{\bmw}~ \mathcal{L}_{\set{D}}(\bmw).
\end{align}

Problem \eqref{eq:erm} cannot be directly solved using standard gradient-based methods since: {\em (i)} the spiking mechanism \eqref{eq:ind-threshold} is not differentiable in $\bmw$ due to the presence of the threshold function $\Theta(\cdot)$; and {\em (ii)} in the case of binary SNNs, the domain of the weight vector $\bmw$ is the discrete set of binary values. 

To tackle the former problem, as detailed in Sec.~\ref{sec:sg}, surrogate gradients (SG) methods replace the derivative of the threshold function $\Theta(\cdot)$ in \eqref{eq:ind-threshold} with a suitable differentiable approximation \cite{neftci2019surrogate}. In a similar manner, for the latter issue, optimization over binary weights is conventionally done via the straight-through estimator (STE) \cite{bengio2013ste, jang21bisnn}, which is covered in Sec.~\ref{sec:ste}. 

\subsubsection{Surrogate Gradient}
\label{sec:sg}
As discussed in the previous subsections, the gradient $\nabla_{\bmw} \mathcal{L}_{\bm{x}, \bm{y}}(\bmw)$ is typically evaluated via SG methods. SG techniques approximate the Heaviside function $\Theta(\cdot)$ in \eqref{eq:ind-threshold} when computing the gradient $\nabla_{\bmw}\set{L}_{\bm{x}, \bm{y}}(\bmw)$. Specifically, the derivative  $\Theta'(\cdot)$ is replaced with the derivative of a differentiable surrogate function, such as rectifier or sigmoid. For example, with a sigmoid surrogate, given by function $\sigma(x) = (1+e^{-x})^{-1}$, we have $\partial s_{i,t}/\partial u_{i,t} \approx \sigma'(u_{i,t} - \vartheta)$, with derivative $\sigma'(x) = \sigma(x)(1-\sigma(x))$.  Using the loss decomposition in \eqref{eq:loss-decomp}, the partial derivative of the training loss  $\set{L}_{\bm{x}^t, \bm{y}_t}(\bmw)$ at each time instant $t$ with respect to a synaptic weight $w_{ij}$ can be accordingly approximated as  

\begin{align} \label{eq:loss-der}
    \frac{\partial \set{L}_{\bm{x}^t, \bm{y}_t}(\bmw)}{\partial w_{ij}} \approx \underbrace{ \sum_{m \in \set{Y}} \frac{\partial L(y_{m,t}, f_{m,t})}{\partial s_{i,t}}}_{e_{i,t}} \cdot \underbrace{ \frac{\partial s_{i,t}}{u_{i,t}}}_{\sigma'(u_{i,t}-\vartheta)} \cdot \underbrace{ \frac{\partial u_{i,t}}{\partial w_{ij}}}_{\alpha_t \ast s_{j,t}},
\end{align}
where the first term $e_{i,t}$ is the derivative of the loss at time $t$ with respect to the output $s_{i,t}$ of post-synaptic neuron $i$ at time $t$; and the third term can be directly computed from \eqref{eq:ind-membrane} as the filtered pre-synaptic trace of neuron $j$. For simplicity of notation, we have defined $f_{m,t}: = f_{m}(\bmw, \bmx^t)$ and omitted the explicit dependence of $s_{i,t}$ and $u_{i,t}$ on exogeneous inputs $\bmx^t$ and synaptic weights $\bmw$. The second term is the source of the approximation, as the derivative of the threshold function $\Theta'(\cdot)$ from \eqref{eq:ind-threshold}, which is zero almost everywhere, is replaced using the derivative of the sigmoid function.

At every time instant $t=1,\dots,T$, using \eqref{eq:loss-der}, the online update is obtained via stochastic gradient descent (SGD) as

\begin{align}
    \label{eq:sg-sgd}
    w_{ij, t+1} \leftarrow  w_{ij, t} - \eta \cdot \frac{1}{|\mathcal{B}|} \sum_{(\bm{x}, \bm{y}) \in \mathcal{B}} \frac{\partial \set{L}_{\bm{x}^t, \bm{y}_t}(\bmw_t)}{\partial w_{ij}},
\end{align}
where $\eta > 0$ is a learning rate, and $\set{B} \subseteq \mathcal{D}$ is a mini-batch of examples $(\bmx, \bmy)$ from the training dataset. Note that the sequential implementation of the update \eqref{eq:sg-sgd} over time $t$ requires running a number of copies of the SNN model equal to the size of the mini-batch $\set{B}$. In fact, each input $\bmx$,  with $(\bmx, \bmy) \in \set{B}$, generally causes the spiking neurons to follow distinct trajectories in the space of the membrane potentials. Henceforth, when referring to online learning rules, we will implicitly assume that parallel executions of the SNN are possible when the mini-batch size is larger than $1$.

The weight update in the direction of the negative gradients in \eqref{eq:sg-sgd} implements a standard {\em three-factor} rule. Three-factor rules generalize two-factor Hebbian updates such as STDP \cite{gerstner18neohebbian}, and can be implemented on hardware with similar complexity \cite{zenke2018superspike, kaiser2020decolle, stewart21onchip}. In fact, the partial derivative \eqref{eq:loss-der} can be written as

\begin{align} \label{eq:loss-sg-der}
    \frac{\partial \set{L}_{\bm{x}^t, \bm{y}_t}(\bmw)}{\partial w_{ij}} = \underbrace{e_{i,t}}_{\text{error signal}} \cdot \underbrace{\sigma'(u_{i,t} - \vartheta)}_{\text{post}_{i,t}} \cdot \underbrace{\big( \alpha_t \ast s_{j,t} \big) }_{\text{pre}_{j,t}},
\end{align}
where we distinguish three terms. The first is the per-neuron error signal $e_{i,t}$, which can be in principle computed via backpropagation through time \cite{huh18bptt}. In practice, this term is approximated, e.g. via local signals \cite{bellec20eprop}, or via random projections \cite{kaiser2020decolle}. The latter technique has previously been likened to the biological mechanisms behind short-term memory \cite{zou22spikehd}. We will discuss a specific implementation in Sec.~\ref{sec:impl}. The second contribution is given by the local post-synaptic term $\sigma'(u_{i,t}-\vartheta)$, which measures the current sensitivity to changes in the membrane potential of the neuron $i$. Finally, the last term is the local pre-synaptic trace $\alpha_t \ast s_{j,t}$ that depends on the activity of the neuron $j$.

\subsubsection{Straight-Through Estimator}
\label{sec:ste}
As mentioned in Sec.~\ref{sec:erm}, optimization over binary weights can be carried out using STE \cite{bengio2013ste, jang21bisnn}, which maintains latent, real-valued weights to compute gradients during training. Binary weights, obtained via quantization of the real-valued latent weights, are used as the next iterate. 
To elaborate, in addition to the binary weight vector $\bmw \in \{+1,-1\}^{|\bmw|}$, we define the real-valued weight vector $\bmw^\text{r} \in \mathbb{R}^{|\bmw| \times 1}$. We use $|\bmw|$ to denote the size of vector $\bmw$.  With STE, gradients are estimated by differentiating over the real-valued latent weights $\bmw^\text{r}$, instead of discrete binary weights $\bmw$, to compute the gradient $\nabla_{\bmw^{\text{r}}}\set{L}_{\bm{x}^t, \bm{y}_t}(\bmw^{\text{r}})|_{\bmw^{\text{r}} = \bmw}$. The technique can be naturally combined with the SG method, detailed in Sec.~\ref{sec:sg}, to obtain the gradients with respect to the real-valued latent weights. 

The training algorithm proceeds iteratively by selecting a mini-batch $\set{B}$ of examples $(\bmx, \bmy)$ from the training dataset $\set{D}$ at each iteration as in \eqref{eq:sg-sgd}. Accordingly, the real-valued latent weight vector $\bmw^{\text{r}}$ is updated via online SGD as

\begin{align} \label{eq:st-update}
   w_{ij, t+1}^\text{r} \leftarrow w_{ij, t}^\text{r} - \eta \cdot \frac{1}{|\set{B}|} \sum_{(\bmx, \bmy) \in \set{B}}  \frac{\partial \set{L}_{\bm{x}^t, \bm{y}_t}(\bmw^{\text{r}}_t)}{\partial w_{ij,t}^{\text{r}}}\bigg|_{w_{ij,t}^{\text{r}} = w_{ij,t}},
\end{align}

and the next iterate for the binary weights $\bmw$ is obtained by quantization as

   \begin{align} \label{eq:st-binarize}
       w_{ij, t+1} = \text{sign}\big(w_{ij, t+1}^{\text{r}}\big),
   \end{align}
   
where the sign function is defined as $\text{sign}(x) = +1$ for $x \geq 0$ and $\text{sign}(x) = -1$ for $x < 0$.





\subsection{Bayesian Offline Learning}
\label{sec:bayes_snn}
In this section, we describe the formulation of Bayesian offline learning, and then develop two Bayesian training algorithms for SNNs with real-valued and binary synaptic weights.

\subsubsection{Information Risk Minimization}
\label{sec:irm}
Bayesian learning formulates the training problem as the optimization of a probability distribution $q(\bmw)$ in the space of synaptic weights, which is referred to as the {\em variational posterior}. To this end, the ERM problem \eqref{eq:erm} is replaced by the information risk minimization (IRM) problem

\begin{align} \label{eq:irm}
    \min_{q(\bmw)}~ \Big\{ \set{F}\big(q(\bmw)\big) = \mathbb{E}_{q(\bmw)} \Big[ \mathcal{L}_{\set{D}}(\bmw) \Big] + \rho \cdot \text{KL}\big(q(\bmw) || p(\bmw) \big) \Big\},
\end{align}

where $\rho > 0$ is a ``temperature'' constant, $p(\bmw)$ is an arbitrary prior distribution over synaptic weights, and $\text{KL}(\cdot||\cdot)$ is the Kullback-Leibler divergence

\begin{align}
    \text{KL}(q(\bmw)||p(\bmw)) = \mathbb{E}_{q(\bmw)}\Bigg[ \log \frac{q(\bmw)}{p(\bmw)}\Bigg].
\end{align}

The objective function in IRM problem \eqref{eq:irm} is known as (variational) free energy \cite{jose20free}.

The problem of minimizing the free energy in \eqref{eq:irm} must strike a balance between fitting the data -- i.e., minimizing the first term -- and not deviating too much from the reference behavior defined by prior $p(\bmw)$ -- i.e., keeping the second term small. Note that with $\rho = 0$, the IRM problem \eqref{eq:irm} reduces to the ERM problem \eqref{eq:erm} in the sense that the optimal solution of the IRM problem with $\rho = 0$ is a distribution concentrated at the solution of the ERM problem (assuming that the latter is unique). The KL divergence term in \eqref{eq:irm} is hence essential to Bayesian learning, and it is formally justified as a regularizing penalty that accounts for epistemic uncertainty due to the presence of limited data in the context of PAC Bayes theory \cite{zhang2006itbound}. It can also be used as a model of bounded rationality accounting for the complexity of information processing \cite{jose20free}.

If no constraints are imposed on the variational posterior $q(\bmw)$, the optimal solution of \eqref{eq:irm} is given by the {\em Gibbs posterior}

\begin{align} \label{eq:bayes-sol}
    q^\star(\bmw) = \frac{ p(\bmw) \exp\big( - \mathcal{L}_{\mathcal{D}}(\bmw) / \rho \big)}{\mathbb{E}_{p(\bmw)} \Big[ \exp\big( - \mathcal{L}_{\mathcal{D}}(\bmw) / \rho \big) \Big]}.
\end{align}

Due to the intractability of the normalizing constant in \eqref{eq:bayes-sol}, we adopt a mean-field variational inference (VI) approximation that limits the optimization domain for problem \eqref{eq:irm} to a class of factorized distributions (see, e.g., \cite{angelino16bayesian, simeone22ml4eng}). More specifically, we focus on Gaussian and Bernoulli variational approximations, targeting SNN models with real-valued and binary synaptic weights, respectively, which are detailed in the rest of this section.

\subsubsection{Gaussian Mean-Field Variational Inference}
\label{sec:bayes-gausssnn}

\DecMargin{2em}
\begin{algorithm}[t]
\caption{Bayesian offline learning with real-valued synapses}
\label{alg:bayes-gaussnn}
\begin{algorithmic}[1]
   \STATE {\bfseries Input:} dataset $\mathcal{D}$, learning rate $\eta$, temperature parameter $\rho$, prior $(\bm{m}_0, \bm{p}_0)$
   
   \STATE{\bfseries Output:} learned parameters pair $(\bm{m}, \bm{p})$
   \STATE {\bf initialize} parameters $(\bm{m}_1, \bm{p}_1)$
   \REPEAT
   \STATE select mini-batch $\mathcal{B} \subseteq \mathcal{D}$
  \smallskip
   \FOR{each time-step $t=1, \dots, T$}
   \smallskip
   \STATE sample weights $\bmw$ as $\bmw \sim \mathcal{N}(\bmw | \bm{m}_t, \bm{P}^{-1}_t)$.
   \smallskip
   \FOR{each $(\bm{x}, \bm{y}) \in \mathcal{B}$}
   \STATE compute the gradient  $\nabla_{\bm{w}} \set{L}_{\bm{x}^t, \bm{y}_t}(\bmw)$ locally at each synapse using SG (see Sec.~\ref{sec:sg}).
   \ENDFOR
   \STATE update the mean and precision parameters $(m_{ij,t}, p_{ij,t})$ for all synapses $(i,j) \in \set{E}$ as
    \begin{align*}
     & p_{ij, t+1} \leftarrow (1 - \eta \rho) \cdot p_{ij,t} + \eta \cdot \bigg[ \frac{1}{|\mathcal{B}|} \sum_{(\bm{x}, \bm{y}) \in \mathcal{B}} \bigg(\frac{\partial\set{L}_{\bm{x}^t, \bm{y}_t}(\bmw)}{\partial w_{ij}} \bigg)^2  +\rho \cdot p_{ij, 0} \bigg] & \\
     & m_{ij, t+1} \leftarrow  m_{ij, t} - \eta \cdot p^{-1}_{ij, t+1} \cdot \bigg[ \frac{1}{|\mathcal{B}|} \sum_{(\bm{x}, \bm{y}) \in \mathcal{B}} \frac{\partial \set{L}_{\bm{x}^t, \bm{y}_t}(\bmw)}{\partial w_{ij}} - \rho \cdot p_{ij, 0} \cdot \big(m_{ij, 0} - m_{ij, t}\big) \bigg]. &
    \end{align*}
    \ENDFOR
    \STATE set $(\bm{m}_1, \bm{p}_1) = (\bm{m}_T, \bm{p}_T)$
   \UNTIL{convergence}
\end{algorithmic}
\end{algorithm}
\IncMargin{2em}

In this subsection, we derive a Gaussian mean-field VI algorithm that approximately solves the IRM problem \eqref{eq:irm} by assuming variational posteriors of the form $q(\bmw) =  \mathcal{N}(\bmw | \bm{m}, \bm{P}^{-1})$, where $\bm{m}$ is a mean vector and $\bm{P}$ is a precision diagonal matrix with positive vector ${\bm p}$ on the main diagonal. For the $|\bmw|\times1$ weight vector $\bmw$, the distribution of the parameters $\bmw$ is defined by the $|\bmw|\times 1$ mean vector $\bm{m}$ and $|\bmw| \times 1$ precision vector $\bm{p} = \{p_{ij}\}_{(i, j) \in \set{E}}$ with $p_{ij} > 0$ for all $(i, j) \in \set{E}$. 
This variational model is well suited for real-valued synapses, which can be practically realized to the fixed precision allowed by the hardware implementation \cite{davies2018loihi}. We choose the prior $p(\bmw)$ as the Gaussian distribution $p(\bmw) = \mathcal{N}(\bmw | \bm{m}_0, \bm{P}_0^{-1})$ with mean $\bm{m}_0$ and precision matrix $\bm{P}_0$ with positive diagonal vector $\bm{p}_0$.

We tackle the IRM problem \eqref{eq:irm} with respect to the so-called {\em variational parameters} $(\bm{m}, \bm{p})$ of the Gaussian variational posterior $q(\bmw)$ via the Bayesian learning rule \cite{khan21bayesian}. The Bayesian learning rule is derived by applying {\em natural gradient descent} to the variational free energy $\mathcal{F}(q(\bmw))$ in \eqref{eq:irm}. The derivation leverages the fact that the distribution $q(\bmw)$ is an exponential-family distribution with natural parameters $\bm{\lambda} = (\bm{P}\bm{m}, -1/2\bm{P})$, sufficient statistics $\mathbf{T} = (\bmw, \bmw\bmw^T)$ and mean parameters $\mathbf{\bm{\mu}} = (\bm{m}, \bm{P}^{-1} + \bm{m}\bm{m}^{T})$. Updates to the mean $\bm{m}_t$ and precision $\bm{p}_t$ parameters at iteration $t$ can be obtained as \cite{osawa2019practical,khan21bayesian}

\begin{align} 
    \label{eq:bayes-update-mean-scalar-ogn}
     & p_{ij, t+1} \leftarrow (1 - \eta \rho) \cdot p_{ij,t} + \eta \cdot \mathbb{E}_{q_t(\bmw)} \bigg[ \frac{1}{|\mathcal{B}|} \sum_{(\bmx, \bmy) \in \mathcal{B}} \bigg(\frac{\partial\set{L}_{\bm{x}^t, \bm{y}_t}(\bmw)}{\partial w_{ij}} \bigg)^2  +\rho \cdot p_{ij, 0} \bigg] & \\
     \label{eq:bayes-update-prec-scalar-ogn}
     & m_{ij, t+1} \leftarrow  m_{ij, t} - \eta \cdot p^{-1}_{ij, t+1} \cdot \mathbb{E}_{q_t(\bmw)}\bigg[ \frac{1}{|\mathcal{B}|} \sum_{(\bmx, \bmy) \in \mathcal{B}} \frac{\partial \set{L}_{\bm{x}^t, \bm{y}_t}(\bmw)}{\partial w_{ij}} - \rho \cdot p_{ij, 0} \cdot \big(m_{ij, 0} - m_{ij, t}\big) \bigg] &
\end{align}
where where $\eta > 0$ is a learning rate; $\mathcal{B} \subseteq \mathcal{D}$ is a mini-batch of examples $(\bmx, \bmy)$ from the training dataset; and $q_t(\bmw) = \mathcal{N}(\bmw | \bm{m}_t, \bm{P}_t^{-1})$ is the variational posterior at iteration $t$ with $\bm{m}_t$ and $\bm{p}_t$.

In practice, the updates \eqref{eq:bayes-update-mean-scalar-ogn}-\eqref{eq:bayes-update-prec-scalar-ogn} are estimated by evaluating the expectation over distribution $q_t(\bmw)$ via one or more randomly drawn samples $\bmw \sim q_t(\bmw)$. Furthermore, the gradients $\nabla_{\bmw} \set{L}_{\bm{x}^t, \bm{y}_t}(\bmw)$ can be approximated using the online SG method described in Sec.~\ref{sec:sg}. The overall training algorithm proceeds iteratively by selecting a mini-batch $\mathcal{B}  \subseteq \mathcal{D}$ of examples $(\bm{x}, \bm{y})$ from the training dataset at each iteration, and is summarized in Algorithm~\ref{alg:bayes-gaussnn}. Note that, as mentioned in Sec.~\ref{sec:sg}, the implementation of a rule operating with mini-batches requires running $|\set{B}|$ SNN models in parallel, where $|\set{B}|$ is the cardinality of the mini-batch. When this is not possible, the rule can be applied with mini-batches of size $|\set{B}| = 1$.

\subsubsection{Bernoulli Mean-Field Variational Inference} 
\label{sec:bayes-bisnn}
In this subsection, we turn to the case of binary synaptic weights $w_{ij} \in \{+1,-1\}$. For this setting, we adopt the variational posterior $q(\bm{w}) = \text{Bern}\big( \bm{w}|\bm{p} \big)$, with

\begin{align} \label{eq:exp-bernoulli}
    q(\bm{w}) = \prod_{i \in \set{N}} \prod_{j \in \mathcal{P}_i} p_{ij}^{\frac{1+w_{ij}}{2}} (1-p_{ij})^{\frac{1-w_{ij}}{2}},
\end{align}

where the $|\bmw| \times 1$ vector $\bm{p} = \{\{p_{ij}\}_{j \in \mathcal{P}_i}\}_{i \in \set{N}}$ defines the variational posterior, with $p_{ij}$ being the probability that synaptic weights $w_{ij}$ equals $+1$.

The variational posterior \eqref{eq:exp-bernoulli} can be reparameterized in terms of the mean parameters $\bm{\mu} = \{\{\mu_{ij}\}_{j \in \mathcal{P}_i}\}_{i \in \set{N}}$ as

\begin{align}
q(\bm{w}) = \text{Bern}\Big( \bm{w} \Big| \frac{\bm{\mu}+\bm{1}}{2} \Big)    
\end{align}

by setting $p_{ij} = (\mu_{ij}+1)/2$, where $\bm{1}$ is the all-ones vector. It can also be expressed in terms of the logits, or natural parameters, $\bm{w}^{\text{r}} = \{\{w_{ij}^\text{r}\}_{j \in \mathcal{P}_i}\}_{i \in \set{N}}$ as $q(\bm{w}) = \text{Bern}\big( \bm{w} | \sigma(2\bm{w}^{\text{r}}) \big)$ by setting

\begin{align} \label{eq:bernoulli-param}
    w_{ij}^{\text{r}} = \frac{1}{2} \log \bigg(\frac{p_{ij}}{1-p_{ij}}\bigg) = \frac{1}{2} \log \bigg( \frac{1+\mu_{ij}}{1-\mu_{ij}} \bigg),
\end{align}

for all $(i, j) \in \set{E}$. The notation $\bm{w}^{\text{r}}$ has been introduced to suggest a relationship with the STE method described in Sec.~\ref{sec:ste}, as defined below. We assume that the prior distribution $p(\bm{w})$ also follows a mean-field Bernoulli distribution of the form $p(\bm{w}) = \text{Bern}(\bm{w} | \sigma(2\bm{w}_{0}^{\text{r}}))$, for some vector of $\bm{w}_{0}^{\text{r}}$ logits. For example, setting $\bm{w}_{0}^{\text{r}} = {\bm 0}$ indicates that the binary weights are equally likely to be either $+1$ or $-1$ a priori.

\DecMargin{2em}
\begin{algorithm}[t]
\caption{Bayesian offline learning with binary-valued synapses}
\label{alg:bayes-bisnn}
\begin{algorithmic}[1]
   \STATE {\bfseries Input:} dataset $\mathcal{D}$, learning rate $\eta$, temperature parameter $\rho$, GS trick parameter $\tau$, logits $\bm{w}_{0}^{\text{r}}$ of prior distribution 
   
   \STATE{\bfseries Output:} learned real-valued weights $\bm{w}^{\text{r}}$
   
   \vspace{-0.2cm}
   \hrulefill
   
   \STATE {\bf initialize} real-valued weights $\bm{w}_{1}^{\text{r}}$
   \REPEAT
   \STATE select mini-batch $\mathcal{B} \subseteq \mathcal{D}$
  \smallskip
  
  \FOR{each time-step $t=1, \dots, T$}
   \smallskip
   \STATE sample relaxed binary weights as
   \begin{align*}
       w_{ij} = \text{tanh}\bigg( \frac{w_{ij, t}^{\text{r}}+\delta_{ij}}{\tau}\bigg),
   \end{align*}
   with $\delta_{ij} = \frac{1}{2} \log \frac{\epsilon_{ij}}{1-\epsilon_{ij}}$ and $\epsilon_{ij} \stackrel{\text{i.i.d.}}{\sim} \mathcal{U}(0,1)$ for all $(i,j) \in \set{E}$.
   \smallskip
  \FOR{each $(\bm{x}, \bm{y}) \in \mathcal{B}$}
  \STATE compute the gradient $\nabla_{\bm{w}} \set{L}_{\bm{x}^t, \bm{y}_t}(\bmw)$ locally at each synapse using SG (see Sec.~\ref{sec:sg}).
  \ENDFOR
   \STATE update the real-valued weights $w_{ij,t}^{\text{r}}$ for all synapses $(i,j) \in \set{E}$ as
   \begin{align*} 
       w_{ij, t+1}^{\text{r}} \leftarrow (1-\eta \rho) \cdot w_{ij, t}^{\text{r}} - \eta \cdot \Big[ \frac{1-w_{ij}^2}{\tau \big( 1 - \text{tanh}^2(w_{ij,t}^{\text{r}}) \big)}
    \cdot \frac{1}{|\mathcal{B}|} \sum_{(\bm{x}, \bm{y}) \in \mathcal{B}} \frac{\partial \set{L}_{\bm{x}^t, \bm{y}_t}(\bmw)}{\partial w_{ij}} - \rho \cdot w_{ij,0}^{\text{r}} \Big].
   \end{align*}
  \ENDFOR
  \STATE set $\bmw_{1}^{\text{r}} = \bmw_{T}^{\text{r}} $
  \UNTIL{convergence}
\end{algorithmic}
\end{algorithm}
\IncMargin{2em}

In a manner similar to the case of Gaussian VI developed in the previous subsection, we apply natural gradient descent to minimize the variational free energy in \eqref{eq:irm} with respect to the variational parameters $\bm{w}^{\text{r}}$ defining the variational posterior $q(\bm{w})$. Following \cite{meng2020bayesbinn}, and applying the online SGD rule detailed in Sec.~\ref{sec:sg}, this yields the update 

\begin{align} \label{eq:bayes-update}
    w_{ij, t+1}^{\text{r}} \leftarrow (1-\eta \rho) \cdot w_{ij, t}^{\text{r}} - \eta \cdot \bigg[ \frac{\partial}{\partial \mu_{ij,t}}\mathbb{E}_{q_t(\bm{w})} \Big[ \frac{1}{|\mathcal{B}|} \sum_{(\bm{x}, \bm{y}) \in \mathcal{B}} \set{L}_{\bm{x}^t, \bm{y}_t}(\bmw) \Big] - \rho \cdot w_{ij, 0}^{\text{r}} \bigg],
\end{align}

where $\eta > 0 $ is a learning rate and $q_t(\bmw)$ the variational posterior with $\bmw^\text{r}_t$ and $\bm{\mu}_t$  related through \eqref{eq:bernoulli-param}. Note that the gradient in \eqref{eq:bayes-update} is with respect to the mean parameters $\bm{\mu}_t$. 

In order to estimate the gradient in \eqref{eq:bayes-update}, we leverage the {\em reparameterization} trick via the {\em Gumbel-Softmax} (GS) distribution \cite{jang2016categorical, meng2020bayesbinn}. Accordingly, we first obtain a sample $\bm{w}$ that is approximately distributed according to $q_t(\bm{w}) = \text{Bern}\big( \bm{w} | \sigma(2\bm{w}^{\text{r}}_t) \big)$. This is done by drawing a vector $\bm{\delta} = \{\{\delta_{ij}\}_{j \in \mathcal{P}_i}\}_{i \in \set{N}}$ of i.i.d. Gumbel variables, and computing 

\begin{align} \label{eq:GS-trick}
    w_{ij} = \text{tanh}\bigg( \frac{w^{\text{r}}_{ij,t} + \delta_{ij}}{\tau} \bigg),
\end{align}

where $\tau > 0$ is a parameter. When $\tau$ in \eqref{eq:GS-trick} tends to zero, the $\text{tanh}(\cdot)$ function tends to the $\text{sign}(\cdot)$ function, and the vector $\bm{w}$ follows distribution $q_t(\bm{w})$ \cite{meng2020bayesbinn}. To generate $\bm{\delta}$, one can set $\delta_{ij} = \frac{1}{2} \log \left(\frac{\epsilon_{ij}}{1-\epsilon_{ij}}\right)$, with $\epsilon_{ij} \sim \mathcal{U}(0,1)$ being i.i.d. samples.

With this sample, for each example $(\bm{x}, \bm{y})$, we then obtain an approximately unbiased estimate of the gradient in \eqref{eq:bayes-update} by using the following approximation 

\begin{align} \label{eq:bayes-RP-grad}
    \frac{\partial}{\partial \mu_{ij,t}} \mathbb{E}_{q_t(\bm{w})} \Big[ \set{L}_{\bm{x}^t, \bm{y}_t}(\bmw) \Big] & \stackrel{(a)}{\approx} \mathbb{E}_{p(\bm{\delta})} \bigg[ \frac{\partial \set{L}_{\bm{x}^t, \bm{y}_t}(\bmw)}{\partial \mu_{ij,t}} \bigg|_{\bmw = \text{tanh}\big(\frac{\bm{w}_{t}^{\text{r}}+\bm{\delta}}{\tau} \big)}\bigg] & \cr 
    & \stackrel{(b)}{=} \mathbb{E}_{p(\bm{\delta})} \bigg[ \frac{\partial \set{L}_{\bm{x}^t, \bm{y}_t}(\bmw)}{\partial w_{ij}} \cdot \frac{\partial}{\partial \mu_{ij,t}} \text{tanh}\Big( \frac{w^{\text{r}}_{ij,t}+\delta_{ij}}{\tau} \Big) \bigg] & \cr 
    & = \mathbb{E}_{p(\bm{\delta})} \bigg[ \frac{\partial \set{L}_{\bm{x}^t, \bm{y}_t}(\bmw)}{\partial w_{ij}} \cdot \frac{1-w^2_{ij}}{\tau \big( 1 - \text{tanh}^2(w^{\text{r}}_{ij,t}) \big)} \bigg], &
\end{align}

where the approximate equality (a) is exact when $\tau \rightarrow 0$ and the equality (b) follows the chain rule. We note that the gradient $\nabla_{\bm{w}} \set{L}_{\bm{x}^t, \bm{y}_t}(\bmw)$ can be computed as detailed in Sec.~\ref{sec:sg}.

As summarized in Algorithm~\ref{alg:bayes-bisnn}, the resulting rule proceeds iteratively by selecting a mini-batch $\mathcal{B}$ of examples $(\bm{x}, \bm{y})$ from the training dataset $\mathcal{D}$ at each iteration. Using the samples $w_{ij}$ from \eqref{eq:GS-trick}, we obtain at every time-step $t$ the estimate of the gradient \eqref{eq:bayes-update} as

\begin{align} \label{eq:bayes-RP-grad-estimator}
    \frac{\partial}{\partial \mu_{ij,t}} \mathbb{E}_{q_t(\bm{w})} \Big[ \set{L}_{\bm{x}^t, \bm{y}_t}(\bmw) \Big] \approx \frac{1-w^2_{ij}}{\tau \big( 1 - \text{tanh}^2(w^{\text{r}}_{ij,t}) \big)}
    \cdot \frac{1}{|\mathcal{B}|} \sum_{(\bm{x}, \bm{y}) \in \mathcal{B}}  \frac{\partial \set{L}_{\bm{x}^t, \bm{y}_t}(\bmw)}{\partial w_{ij}} - \rho \cdot w^\text{r}_{ij,0}.
\end{align}
This is unbiased when the limit $\tau \rightarrow 0$ holds.





\subsection{Frequentist Continual Learning}
\label{sec:freq-online}
We now consider a continual learning setting, in which the system is sequentially presented datasets $\mathcal{D}^{(1)}, \mathcal{D}^{(2)}, \dots$ corresponding to distinct, but related, learning tasks. Applying a frequentist approach, at every subsequent task $k$, the system minimizes a new objective based on dataset $\set{D}^{(k)}$ in order to update the model parameter vector $\bmw$, where we have used superscript $(k)$ to denote the quantities corresponding to the $k$th task. We first describe an algorithm based on coresets and regularization \cite{farquhar19unifying}. Then, we briefly review a recently proposed biologically inspired rule. 

\subsubsection{Regularization-based Continual Learning}
In a similar manner to \eqref{eq:loss}, let us first define as 

\begin{align} \label{eq:continual-freq-loss}
    \mathcal{L}_{\set{D}^{(k)}}(\bmw) = \frac{1}{|\mathcal{D}^{(k)}|} \sum_{(\bm{x}, \bm{y}) \in \mathcal{D}^{(k)}} \mathcal{L}_{\bm{x}, \bm{y}}(\bmw)
\end{align}

the training loss evaluated on dataset  $\set{D}^{(k)}$ for the $k$th task. A general formulation of the continual learning problem in a frequentist framework is then obtained as the minimum of the objective

\begin{align} \label{eq:continual-freq-reg}
    \mathcal{L}_{\set{D}^{(k)}}(\bmw) + \sum_{k'= 1}^{k-1} \mathcal{L}_{\set{C}^{(k')}}(\bmw) + \alpha \cdot R\big(\bmw, \{\bmw^{(k')}\}_{k'=1}^{k-1}\big),
\end{align}

where $\mathcal{L}_{\mathcal{C}^{(k')}}(\bmw)$ is the training loss evaluated on a \textit{coreset}, that is, a subset $\mathcal{C}^{(k')} \subset \mathcal{D}^{(k')}$ of examples randomly selected from a previous task $k'< k$ and maintained for use when future tasks are encountered; $\alpha \geq 0$ determines the strength of the regularization; and $R(\bmw, \{\bmw^{(k')}\}_{k'=1}^{k-1})$ is a regularization function aimed at preventing the current weights from differing too much from previously learned weights $\{\bmw^{(k')}\}_{k'=1}^{k-1}$, hence mitigating the problem of catastrophic forgetting \cite{parisi18continual}. 

A popular choice for the regularization function, yielding the Elastic Weight Consolidation (EWC) method, proposes to estimate the relative importance of synapses for previous tasks via the Fisher information matrices (FIM) computed on datasets $k' < k$ \cite{kirkpatrick17ewc}. This corresponds to the choice of the regularizer

\begin{align} \label{eq:ewc}
     R\big(\bmw, \{\bmw^{(k')}\}_{k'=1}^{k-1}\big) = \sum_{k' = 1}^{k-1} (\bmw - \bmw^{(k')})^T F^{(k')}(\bmw^{(k')}) (\bmw - \bmw^{(k')}),
\end{align}

where $F^{(k)}(\bmw) = \text{diag}\big(\sum_{(\bm{x}, \bm{y}) \in \mathcal{D}^{(k)}} (\nabla_{\bmw}\mathcal{L}_{\bm{x}, \bm{y}}(\bmw))^2 \big)$ is an approximation of the FIM estimated on dataset $\mathcal{D}^{(k)}$. The square operation in vector $(\nabla_{\bmw}\mathcal{L}_{\bm{x}, \bm{y}}(\bmw))^2$ is evaluated element-wise. Intuitively, a larger value of an entry in the diagonal of the matrix $F^{(k)}(\bmw)$ indicates that the corresponding entry of the vector $\bmw$ plays a significant role for the $k$th task.

\subsubsection{Biologically Inspired Continual Learning}
The authors of \cite{soures21tacos} introduce a biologically inspired, frequentist, continual learning rule for SNNs, which we briefly review here. The approach operates online in discrete time $t$, and implements the mechanisms described in Sec.~\ref{sec:biological-learning}. It considers a leaky integrate-and-fire (LIF) neuron model. The LIF is a special case of the SRM \eqref{eq:ind-threshold}-\eqref{eq:ind-membrane} in which the synaptic response $\alpha$ implemented as the \textit{alpha-function} spike response $\alpha_t = \exp(-t/\tau_{\text{mem}}) - \exp(-t/\tau_{\text{syn}})$ and the exponentially decaying feedback filter $\beta_t = -\exp(-t/\tau_{\text{ref}})$ for $t \geq 1$ with some positive constants $\tau_{\text{mem}}, \tau_{\text{syn}}$, and $\tau_{\text{ref}}$. This choice enables an autoregressive update of the membrane potential \cite{kaiser2020decolle, snnreview}. 

A metaplasticity parameter $\nu_{ij}$ is introduced for each synapse $(i, j) \in \set{E}$ that determines the degree to which the synapse is prone to change. This quantity is increased by a fixed step $\Delta \nu$ as

\begin{align}
\nu_{ij, t+1} \leftarrow \nu_{ij,t} + \Delta \nu    
\end{align}

when the pre- and post-synaptic neurons spiking rates, i.e., the spiking rate of neuron $i$ and $j$, respectively, pass a pre-determined threshold. Furthermore, each synapse $(i,j) \in \set{E}$ maintains a reference weight $w^{\text{ref}}_{ij}$ to mimic heterosynaptic plasticity by adjusting the weight updates to drive synaptic weights towards this resting state. It is updated over time as

\begin{align}
    \label{eq:tacos-ref}
    w^{\text{ref}}_{ij, t+1} \leftarrow w^{\text{ref}}_{ij, t} + \kappa \cdot \Big( w_{ij, t} - w^{\text{ref}}_{ij, t} \Big),
\end{align}

where $\kappa > 0$, and serves as a reference to implement heterosynaptic plasticity.

With these definitions, the update of each synaptic weight $\bmw$ is computed according to the online learning rule

\begin{align}
    \label{eq:tacos-lr}
    w_{ij, t+1} \leftarrow w_{ij, t}  - \exp{\big(- |\nu_{ij} \cdot w_{ij,t}|\big)}\Big( \eta \cdot e_{i,t} \cdot s_{j,t} \cdot \sigma'(u_{i,t} - \vartheta) + \gamma \cdot (w_{ij,t} - w^{\text{ref}}_{ij,t})\cdot s_{i,t}\Big),
\end{align}

where $\eta$ and $\gamma$ are respectively learning and decay rates, and $e_{i,t}$ is an error signal from neuron $i$ (see \cite{soures21tacos} for details). The rule \eqref{eq:tacos-lr} takes a form similar to that of three-factor rules \eqref{eq:loss-sg-der}, with the term $e_{i,t} \cdot s_{j,t} \cdot \sigma'(u_{i,t} - \vartheta)$ evaluating the product of error, post-synaptic, and pre-synaptic signals. The update \eqref{eq:tacos-lr} implements metaplasticity via the term $\exp{\big(- |\nu_{ij} \cdot w_{ij,t}|\big)}$ that decreases the magnitude of the updates during the training procedure for active synapses. It also accounts for heterosynaptic plasticity thanks to the term $(w_{ij,t} - w^{\text{ref}}_{ij,t})$, which drives the updates towards learned ``resting'' weight $w^{\text{ref}}_{ij,t}$ when the pre-synaptic neuron is active. 





\subsection{Bayesian Continual Learning}
\label{sec:bayes_cont_snn}
In this section, we generalize the Bayesian formulation seen in Section~\ref{sec:bayes_snn} from the offline setting to continual learning. 

\subsubsection{Bayesian Continual Learning}
To allow the adaptation to task $k$ without catastrophic forgetting, we consider the problem \cite{farquhar19robust} 
\begin{align}
    \min_{q^{(k)}(\bmw)}~ \mathcal{F}^{(k)}\big(q^{(k)}(\bmw)\big)
\end{align}

of minimizing the free energy metric 

\begin{align} 
    \label{continual:eq:bayes-hybrid}
     \mathcal{F}^{(k)}\big(q^{(k)}(\bmw)\big) = \mathbb{E}_{q^{(k)}(\bmw)} \Big[\mathcal{L}_{\set{D}^{(k)}}(\bmw) + \sum_{k' = 1}^{k-1} \mathcal{L}_{\set{C}^{(k')}}(\bmw) \Big] + \rho \cdot \text{KL}\big( q^{(k)}(\bmw) || q^{(k-1)}(\bmw) \big),
\end{align}

which combines the IRM formulation \eqref{eq:irm} with the use of coresets.  Minimizing the free energy objective \eqref{continual:eq:bayes-hybrid} must strike a balance between fitting the new training data $\set{D}^{(k)}$, as well as the coresets $\{\set{C}^{(k')}\}_{k'=1}^{k-1}$ from the previous tasks, while not deviating too much from previously learned distribution $q^{(k-1)}(\bmw)$. Comparing \eqref{continual:eq:bayes-hybrid} with the free energy \eqref{eq:irm}, we observe that the distribution $q^{(k-1)}(\bmw)$ plays the role of prior for the current task $k$.

\subsubsection{Continual Gaussian Mean-Field Variational Inference}
\label{sec:bayes_cont_gausssnn}
Similarly to the approach for offline learning described in Sec.~\ref{sec:bayes_snn}, we first assume a Gaussian variational posterior $q(\bmw)$, and  address the problem \eqref{continual:eq:bayes-hybrid} via natural gradient descent. To this end, we adopt the variational posterior $q^{(k)}(\bmw) = \mathcal{N}(\bmw | \bm{m}^{(k)}, (\bm{P}^{(k)})^{-1})$, with mean vector $\bm{m}^{(k)}$ and diagonal precision matrix $\bm{P}^{(k)}$ with positive diagonal vector $\bm{p}^{(k)}$ of size $|\bmw| \times 1$ for every task $k$. We choose the prior $p(\bmw)$ for dataset $\mathcal{D}_1$ as the Gaussian distribution $p(\bmw) = \mathcal{N}(\bmw | \bm{m}_0, \bm{P}_0^{-1})$ with positive diagonal vector $\bm{p}_0$ of size $|\bmw| \times 1$. Applying the Bayesian learning rule \cite{khan21bayesian} as in Sec.~\ref{sec:bayes-gausssnn}, updates to the mean and precision parameters can be obtained via online SGD as 

\begin{align} 
\label{continual:eq:bayes-update-prec-scalar-final}
   p^{(k)}_{ij, t+1}  &~ \leftarrow (1-\eta \rho) \cdot p^{(k)}_{ij,t} + \eta \cdot \mathbb{E}_{q^{(k)}_t(\bmw)} \bigg[ \frac{1}{|\mathcal{B}|} \sum_{(\bm{x}, \bm{y}) \in \set{B}} \bigg(\frac{\partial\set{L}_{\bm{x}^t, \bm{y}_t}(\bmw)}{\partial w_{ij}} \bigg)^2 + \rho \cdot p^{(k-1)}_{ij} \bigg]& \\ 
   \label{continual:eq:bayes-update-mean-scalar-final}
   m^{(k)}_{ij, t+1}   
   &~\leftarrow m^{(k)}_{ij,t} - \eta \cdot (p^{(k)}_{ij,t+1})^{-1} \cdot \mathbb{E}_{q^{(k)}_t(\bmw)} \bigg[ \frac{1}{|\mathcal{B}|} \sum_{(\bm{x}, \bm{y}) \in \set{B}}\frac{\partial\set{L}_{\bm{x}^t, \bm{y}_t}(\bmw)}{\partial w_{ij}} - \rho \cdot p^{(k-1)}_{ij} \cdot \big( m^{(k-1)}_{ij} - m^{(k)}_{ij,t} \big) \bigg],
\end{align}

where mini-batch $\set{B}$ is now selected at random from both dataset $\set{D}^{(k)}$ and coresets from previous tasks, i.e., $\set{B} \subseteq \set{D}^{(k)} \cup_{k'=1}^{k} \set{C}^{(k')}$. 
The rule can be directly derived by following the steps detailed in Sec.~\ref{sec:bayes-gausssnn}, and using for prior at every task $k$ the mean $\bm{m}^{(k-1)}$ and precision $\bm{P}^{(k-1)}$ obtained at the end of training on the previous task.

\subsubsection{On the Biological Plausibility of the Bayesian Learning Rule}
The continual learning rule \eqref{continual:eq:bayes-update-prec-scalar-final}-\eqref{continual:eq:bayes-update-mean-scalar-final} exhibits some of the mechanisms thought to enable memory retention in biological brains as described in Sec.~\ref{sec:biological-learning}. In particular, synaptic consolidation and metaplasticity for each synapse $(i,j) \in \set{E}$ are modelled by the precision $p_{ij}$. In fact, a larger precision $p_{ij, t+1}$ effectively reduces the step size $1/p_{ij, t+1}$ of the synaptic weight update \eqref{continual:eq:bayes-update-mean-scalar-final}. This is a similar mechanism to the metaplasticity parameter $\nu_{ij,t}$ introduced in the rule \eqref{eq:tacos-lr}. Furthermore, by \eqref{continual:eq:bayes-update-prec-scalar-final}, the precision $p_{ij}$ is increased to a degree that depends on the relevance of the synapse $(i,j) \in \set{E}$ as measured by the estimated FIM $(\partial\set{L}_{\bm{x}^t, \bm{y}_t}(\bmw)/\partial w_{ij})^2$ for the current mini-batch $\set{B}$ of examples.

Heterosynaptic plasticity,  which drives the updates towards previously learned and resting states to prevent catastrophic forgetting, is obtained from first principles via the IRM formulation with a KL regularization term, rather than from the addition of the reference weight $\bmw_{\text{ref}}$ in the previous work \cite{soures21tacos}. This mechanism drives the updates of the precision $p^{(k)}_{ij, t+1}$ and mean parameter $m^{(k)}_{ij, t+1}$ towards the corresponding parameters of the variational posterior obtained at the previous task, namely $p_{ij}^{(k-1)}$ and $m_{ij}^{(k-1)}$.

Finally, the use of coresets implements a form of replay, or reactivation, in biological brains \cite{buhry11replay}. 

\subsubsection{Continual Bernoulli Mean-Field Variational Inference}
\label{sec:bayes_cont_bisnn}

We now consider continual learning with a Bernoulli mean-field variational posterior, and force the synaptic weight $w_{ij}$ to be binary, i.e., $w_{ij} \in \{+1,-1\}$. Following \eqref{eq:exp-bernoulli}, the posterior is of the form $q^{(k)}(\bm{w}) = \text{Bern}\big( \bm{w}|\bm{p}^{(k)} \big)$.

We leverage the Gumbel-softmax trick, and use the reparametrization in terms of the natural parameters at task $k$

\begin{align} \label{eq:cont-bernoulli-param}
    w^{\text{r}, (k)}_{ij} = \frac{1}{2} \log \bigg( \frac{1+\mu^{(k)}_{ij}}{1-\mu^{(k)}_{ij}} \bigg).
\end{align}

We then apply the Bayesian learning rule, and, following the results obtained in the offline learning case of Sec.~\ref{sec:bayes-bisnn}, we obtain the learning rule at task $k$ as

   \begin{align} 
   \label{eq:bayesbisnn-cont}
       w_{ij, t+1}^{\text{r},(k)} \leftarrow (1-\eta \rho) \cdot w_{ij, t}^{\text{r}, (k)} - \eta \cdot \bigg[ \mathbb{E}_{p(\bm{\delta})} \bigg[ \frac{1}{|\mathcal{B}|} \sum_{(\bm{x}, \bm{y}) \in \set{B}} \frac{\partial \set{L}_{\bm{x}^t, \bm{y}_t}(\bmw)}{\partial w_{ij}} \cdot \frac{1-w^2_{ij}}{\tau \big( 1 - \text{tanh}^2 \big(w_{ij, t}^{\text{r},(k)} \big) \big)} \bigg] - \rho \cdot w_{ij}^{\text{r},(k-1)} \bigg],
   \end{align}
   
where we denote as $\bm{w}^{\text{r},(k-1)}$ the logits obtained at the end of the previous task $k-1$, and mini-batch $\set{B}$ is selected at random as $\set{B} \subseteq \set{D}^{(k)} \cup_{k'=1}^{k} \set{C}^{(k')}$.





\section{Experiments}
\label{sec:exp}
In this section, we compare the performance of frequentist and Bayesian learning schemes in a variety of experiments, using both synthetic and real neuromorphic datasets. All experiments consist of classification tasks with $C$ classes. In each such task, we are given a dataset $\set{D}'$ consisting of spiking inputs $\bmx$ and label $c_{\bmx} \in \{0, 1, \dots, C - 1\}$. Each pair $(\bmx, c_{\bmx})$ is converted into a pair of spiking signals $(\bmx, \bmy)$ to obtain the dataset $\set{D}$. To do this, the target signals $\bmy$ are such that each sample $\bmy_t$ is the $C\times 1$ one-hot encoding vector of label $c_{\bmx}$ for all time-steps $t=1, \dots, T$.  

\subsection{Datasets}
\subsubsection{Two-moons dataset}
We first consider an offline 2D binary classification task on the two-moons dataset \cite{twomoonsdataset}. Training is done on $200$ examples per class with added noise with standard deviation $\sigma = 0.1$ as proposed in reference \cite{meng2020bayesbinn} for $100$ epochs. The inputs $\bmx$ are obtained via population encoding following \cite{snnreview} over $T=100$ time-steps and via $10$ neurons. 

\subsubsection{DVS-Gestures}
Next, we consider a real-world neuromorphic dataset for offline classification, namely the DVS-Gestures dataset \cite{amir2017dvsgesture}. The dataset comprises $11$ classes of hand movements, captured with a DVS camera. Movements are recorded from $30$ different persons under $5$ lighting conditions. To evaluate the calibration of Bayesian learning algorithms, we obtain in- and out-of-distribution dataset $\set{D}_{\text{id}}$ and $\set{D}_{\text{ood}}$ by partitioning the dataset by users and lighting conditions. We selected the first $15$ users for the training set, while the remaining $15$ users are used for testing. The first $4$ lighting conditions are used for in-distribution testing; and the one left out from the training set is used for out-of-distribution testing. Images are of size $128 \times 128 \times 2$, and preprocessed following \cite{amir2017dvsgesture} to yield inputs of size $32 \times 32 \times 2$, with sequences of length $500$ ms for training and $1500$ ms for testing, with a sampling rate of $10$ ms. 

\subsubsection{Split-MNIST and MNIST-DVS}
For continual learning, we first conduct experiments on the 5-ways split-MNIST dataset \cite{farquhar19robust, soures21tacos}. Examples from the MNIST dataset, of size $28 \times 28$ pixels, are hence rate-encoded over $T=50$ time-steps \cite{snnreview}, and training examples drawn from subsets of two classes are successively presented to the system for training. The order of the pairs is selected as $\{0, 1\}$, then $\{2, 3\}$, and so on. We restrict here our study to rate encoding, although the proposed methods are applicable to any spike encoding scheme.
In a similar way, we also consider a neuromorphic continual learning setting based on the neuromorphic counterpart to the MNIST dataset, namely, the MNIST-DVS dataset \cite{serrano2015poker}. Following the preprocessing adopted in references \cite{skatchkovsky2020flsnn, skatchkovsky2020neurojscc, skatchkovsky21vdib}, we cropped images spatially to $26 \times 26$ pixels, capturing the active part of the image, and temporally to a duration of $2$ seconds. For each pixel, positive and negative events are encoded as (unsigned) spikes over two different input channels, and the input $\bmx$ is of size $1352$ spiking signals. Uniform downsampling over time is then carried out to restrict the length to $T= 80$ time-samples. The training dataset is composed of $900$ examples per class, and the test dataset contains $100$ examples per class. For continual learning, classes are presented to the network in pairs by following the lexicographical order, i.e., the classes $\{0, 1\}$ are presented first, then $\{2, 3\}$, and so on.

\subsection{Implementation}
\label{sec:impl}
All schemes are implemented using the SG technique DECOLLE \cite{kaiser2020decolle} to compute the gradients. In DECOLLE, the SNN is organized into $L$ layers, with the first $L-1$ layers encompassing the hidden neurons in set $\set{H}$, and the $L$th layer containing the read-out neurons in set $\set{Y}$. To evaluate the partial derivative \eqref{eq:loss-sg-der}, we need to specify how to compute error signals $e_{i,t}$ for each neuron $i \in \set{N}$. To this end, at each time $t$, the spiking outputs $\bms^{(l)}_t$ of each layer $l \in \{1, \dots, L\}$ are used to compute local per-layer errors

\begin{align}
    \label{eq:dcll-loss}
    L(y_{m, t}, \bm{s}^{(l)}_{t}) = - y_{m, t} \cdot \text{log}\Big(\text{Softmax}_m\big(\bm{B}^{(l)}\bms^{(l)}_t\big)\Big),
\end{align}

where $\bm{B}^{(l)} \in \mathbb{R}^{C \times |l|}$ are random, fixed weights, $|l|$ is the cardinality of layer $l$, and $\text{Softmax}_m(\bma) = \text{exp}(a_m) / \sum_{1 \leq n \leq C} \text{exp}(a_n)$ is the $i$th element of the softmax of vector $\bma$ with length $C$. The local losses \eqref{eq:dcll-loss} at every time-step $t$ are then used to compute the error signals $e_{i,t}$  in \eqref{eq:loss-sg-der} for every neuron $i \in l$ as

\begin{align}
    e_{i, t} = \sum_{m \in \set{Y}}\frac{\partial L(y_{m,t}, \bm{s}^{(l)}_{t})}{\partial s_{i,t}}.
\end{align}
While the algorithms introduced in this work are valid for any SNN architecture as highlighted in Fig.~\ref{fig:snn}, DECOLLE is limited to feedforward layered architectures, which we hence adopt for our experiments \cite{kaiser2020decolle}. Furthermore, we consider autoregressive filters for the spike responses of synapses $\alpha_t$ and somas $\beta_t$ in the membrane potential \eqref{eq:ind-membrane}, as discussed in Sec.~\ref{sec:srm}.

Results have been obtained by using Intel's Lava software framework \cite{intel22lava}, under Loihi-compatible fixed-point precision \cite{davies2018loihi}\footnote{Our implementation can be found at \url{https://github.com/kclip/bayesian-snn}.}. We use as benchmark the frequentist algorithms detailed in Sec.~\ref{sec:freq-offline} and \ref{sec:freq-online}, for which gradients are as described in the previous paragraph. For Bayesian learning with real-valued (fixed-precision) synapses, we set the threshold of each neuron as $\vartheta = 64$; while for binary synapses the threshold $\vartheta$ is selected as the square-root of the fan-in of the corresponding layer. 

Implementation of the proposed methods on hardware is left for future work. While Loihi supports the injection of Gaussian noise to the membrane potential of the neurons \cite{davies2018loihi}, it does not provide mechanisms for the sampling of the model parameters. In contrast, recent work \cite{dalgaty21memristor} has proposed leveraging the inherent noise of nanoscale devices in order to implement Bayesian inference.

\subsection{Performance Measures}
\label{sec:perf-meas}
Apart from the test accuracy, performance metrics include calibration measures, namely reliability diagrams and expected calibration error (ECE), which are described next. We note that, as the hardware implementation of Bayesian SNNs is currently an open problem (see Sec.~\ref{sec:impl}), we are unable to provide measurements in terms of energy expenditure and computation time. As a general remark, as discussed in Sec.~\ref{sec:predictors}, Bayesian learning requires a larger memory to store all samples for the weights distribution to be used for inference using a committee machine implementation, while an ensemble predictor implementation increases inference latency.

\subsubsection{Confidence levels}
For frequentist learning, predictive probabilities are obtained from a single pass through the network with parameter vector $\bmw$ as 

\begin{align}
    \label{eq:pred-conf-freq}
    p\big(c_{\bmx} = k \mid  \bmx, \bmw\big) = \frac{1}{T} \sum_{t=1}^{T} \text{Softmax}_k\big(\bm{B}^{(L)}f({\bmw}, \bmx^t)\big),
\end{align}

where $f(\bmw, \bmx^t)$ is the output of read-out layer $L$ for weights $\bmw$, as detailed in the previous subsection.  

In contrast, for Bayesian learning, decisions and confidence levels are obtained by drawing $N_S$ samples $\{\bmw_s \}_{s=1}^{N_S}$ from the distribution $q(\bmw)$, and by averaging the read-out outputs of the model to obtain the probability assigned to each class as

\begin{align}
    \label{eq:pred-conf-bayes}
    p\big(c_{\bmx} = k \mid \bmx, \{\bmw_s \}_{s=1}^{N_S}\big) = \frac{1}{N_S} \frac{1}{T}  \sum_{s=1}^{N_S}\sum_{t=1}^{T} \text{Softmax}_k\big(\bm{B}^{(L)}f({\bmw_s}, \bmx^t)\big).
\end{align}

Unless mentioned otherwise, the predictions \eqref{eq:pred-conf-bayes} are obtained by using the committee machine approach, and hence the weights $\{\bmw_s\}_{s=1}^{N_S}$ are kept fixed for all test inputs $\bmx$ (see Sec.~\ref{sec:predictors}). All results presented are averaged over three repetitions of the experiments and $10$ draws from the posterior distribution $q(\bmw)$, i.e., we set $N_S = 10$ in all experiments. 

For Bayesian learning, the hard prediction of the model is hence obtained as

\begin{align}
    \label{eq:hard-pred}
c_{\bmx}^{*} = \underset{{1\leq k\leq C}}{\text{argmax}}~ p\big(c_{\bmx} = k \mid \bmx, \{\bmw_s \}_{s=1}^{N_S}\big),
\end{align}

corresponding to the predictive probability 

\begin{align}
    \label{eq:prob}
   p\big(c_{\bmx}^{*} \mid \bmx, \{\bmw_s \}_{s=1}^{N_S}\big) = \underset{1\leq k\leq C}{\text{max}}~ p\big(c_{\bmx} = k \mid \bmx, \{\bmw_s \}_{s=1}^{N_S}\big).
\end{align}

The probability \eqref{eq:prob} can be interpreted as the confidence of the model in making decisions \eqref{eq:hard-pred}.

A model is considered to be well calibrated when there is no mismatch between confidence level $p\big(c_{\bmx}^{*} \mid \bmx, \{\bmw_s \}_{s=1}^{N_S}\big)$ and the actual probability for the model to correctly classify input $\bmx$ \cite{guo2017hierarchical}. Definitions \eqref{eq:hard-pred}-\eqref{eq:prob} can be straightforwardly adapted to the frequentist case by replacing the average over draws $\{\bmw_s \}_{s=1}^{N_S}$ with a single parameter vector $\bmw$.

\subsubsection{Reliability diagrams} 

Reliability diagrams plot the actual probability of correct detection as a function of the confidence level \eqref{eq:prob}.  This is done by first dividing the probability interval $[0, 1]$ into $M$ intervals of equal length, and then evaluating the average accuracy and confidence for all inputs $\bmx$ in each $m$th interval $(\frac{m-1}{M}, \frac{m}{M}]$, also referred to as $m$th \textit{bin}. We denote as $\set{B}_m$ the subset of examples whose associated confidence level $p\big(c_{\bmx}^{*} \mid \bmx, \{\bmw_s \}_{s=1}^{N_S}\big)$ lies within bin $m$, that is, \cite{guo2017hierarchical}

\begin{align}
    \set{B}_m = \bigg\{\bmx \in \set{D} ~ \Big| ~ p\big(c_{\bmx}^{*} \mid \bmx, \{\bmw_s \}_{s=1}^{N_S}\big) \in  \Big(\frac{m-1}{M}, \frac{m}{M}\Big] \bigg\}.
\end{align}

The average empirical accuracy of the predictor within bin $m$ is defined as

\begin{figure}[t]
\centering
\includegraphics[width=1.\textwidth]{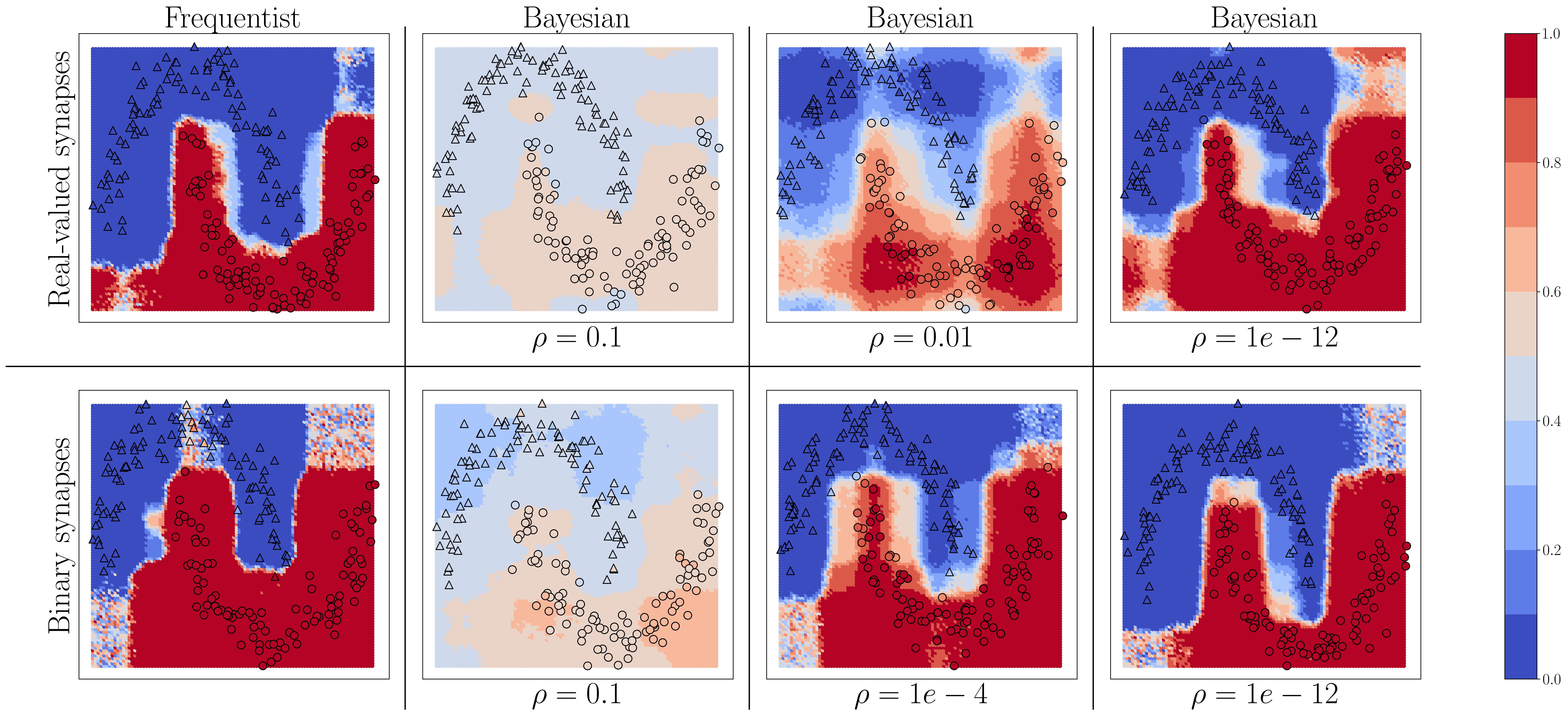} 
\caption{Predictive probabilities \eqref{eq:prob} evaluated on the two-moons dataset after training with different values of the temperature $\rho$ in eq.~\eqref{eq:irm} for Bayesian learning. Top row: Real-valued synapses; Bottom row: Binary synapses.}
\label{fig:twomoons_exp}
\end{figure}

\begin{align}
    \text{acc}(\set{B}_m) = \frac{1}{|\set{B}_m|} \sum_{\bmx \in \set{B}_m} \bm{1}(c_{\bmx}^* = c_{\bmx}),
\end{align}

with $\bm{1}(\cdot)$ being the indicator function; while the average empirical confidence of the predictor for bin $m$ is defined as

\begin{align}
    \text{conf}(\set{B}_m) = \frac{1}{|\set{B}_m|} \sum_{\bmx \in \set{B}_m} p\big(c_{\bmx}^{*} \mid \bmx,  \{\bmw_s \}_{s=1}^{N_S}\big).
\end{align}

\begin{figure}[t]
\centering
\includegraphics[width=\columnwidth]{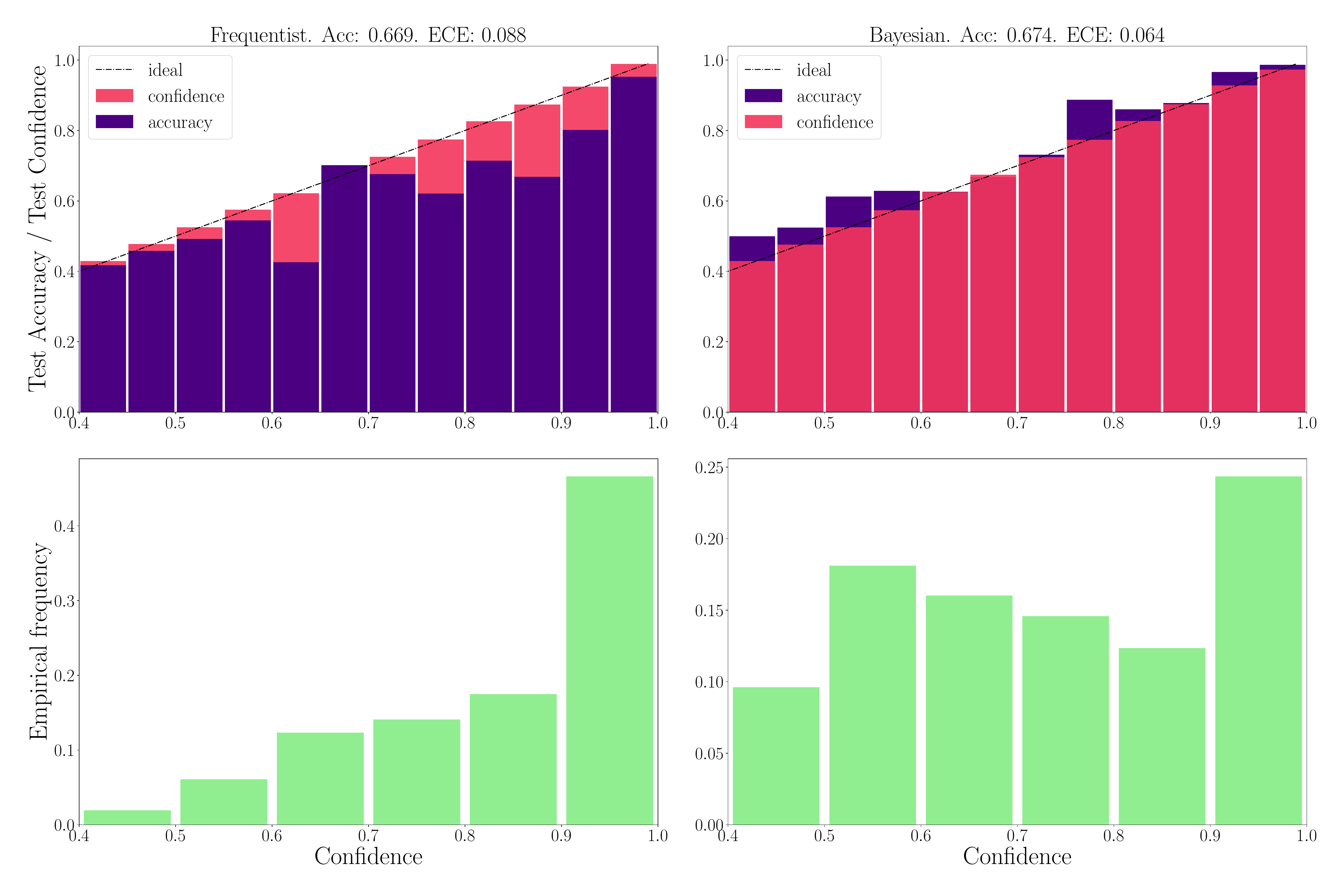}
\caption{Top: Reliability diagrams (for in-distribution data) with real-valued synapses for the DVS-Gestures dataset. Bottom: Corresponding empirical confidence histograms for in-distribution data.}
\label{fig:bayes-id-real}
\end{figure}

Reliability diagrams plot the per-bin accuracy $\text{acc}(\set{B}_m)$ versus confidence level  $\text{conf}(\set{B}_m)$ across all bins $m$. A model is said to be perfectly calibrated when, for all bins $m$, the equality  $\text{acc}(\set{B}_m) = \text{conf}(\set{B}_m)$ holds. If in the $m$th bin, the empirical accuracy
and empirical confidence are different, the predictor is considered to be over-confident when the inequality $\text{acc}(\set{B}_m) < \text{conf}(\set{B}_m)$ holds,
and under-confident when the reverse inequality $\text{acc}(\set{B}_m) > \text{conf}(\set{B}_m)$ holds.

\begin{figure}[t]
\centering
\includegraphics[width=\columnwidth]{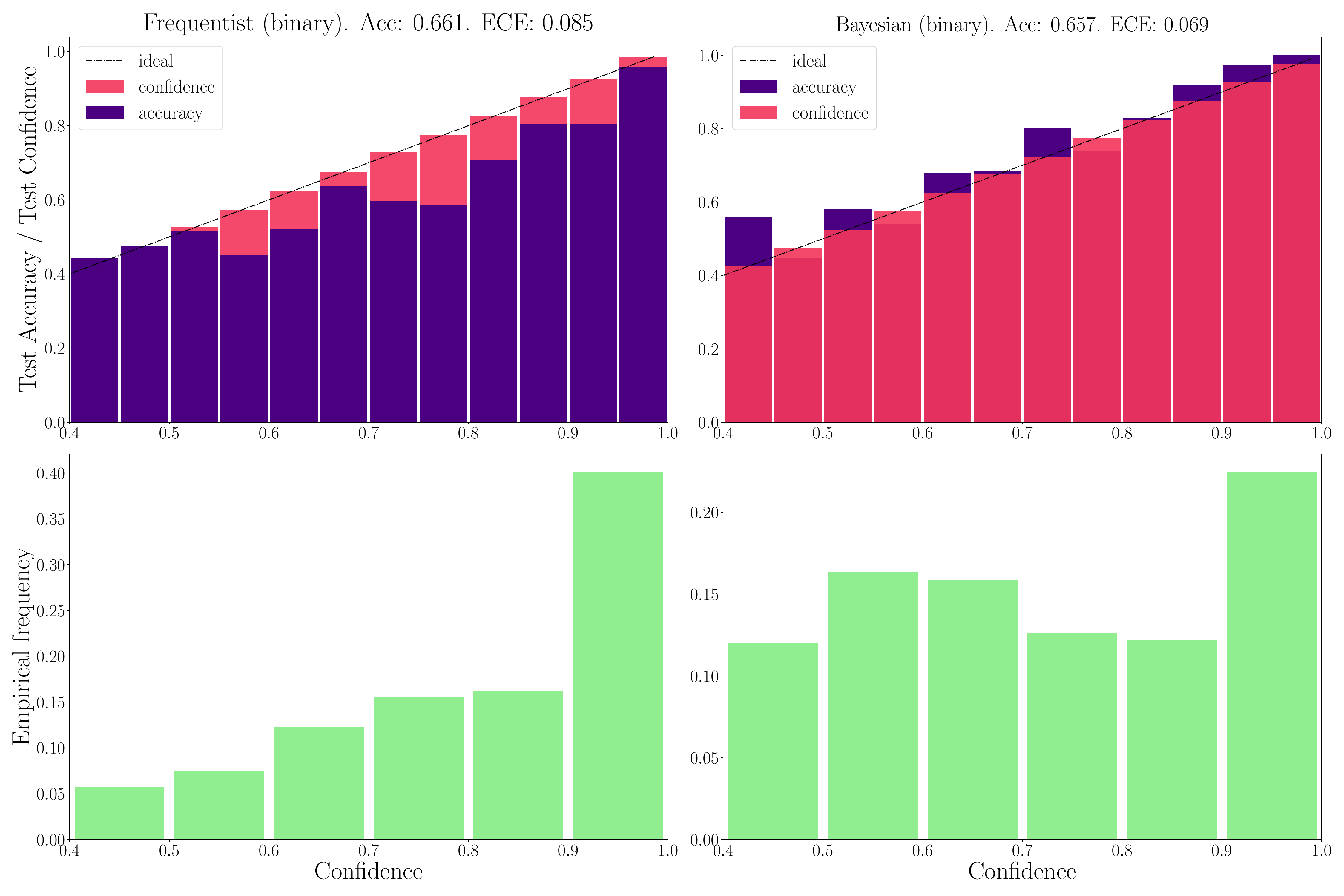}
\caption{Top: Reliability diagrams (for in-distribution data) with binary synapses for the DVS-Gestures dataset. Bottom: Corresponding empirical confidence histograms for in-distribution data.}
\label{fig:bayes-id-bin}
\end{figure}

\subsubsection{Expected calibration error (ECE)}

While reliability diagrams offer a fine-grained description of calibration, the ECE provides a scalar measure of the global miscalibration of the model. This is done by computing the average difference between per-bin confidence and accuracy as \cite{guo2017hierarchical}
\begin{align}
    \label{eq:ece}
    \text{ECE} = \frac{1}{|\set{D}|} \sum_{m=1}^{M} |\set{B}_m| \big| \text{conf}(\set{B}_m) - \text{acc}(\set{B}_m) \big|.
\end{align}
Models with a lower ECE are considered to be better calibrated. 

\subsubsection{Out-of-distribution empirical confidence}

Reliability diagrams and ECE assume that the test data follows the same distribution as the training data. A well-calibrated model is also expected to assign lower probabilities to out-of-distribution data, i.e., data that does not follow the training distribution \cite{degroot83evaluation}. To gauge the capacity of a model to recognize out-of-distribution data, a common approach is to plot the histogram of the predictive probabilities $\big\{p\big(c_{\bmx}^{*} | \bmx, \{\bmw_s \}_{s=1}^{N_S}\big)\big\}_{\bmx \in \set{D}_{\text{ood}}}$ evaluated on a dataset $\set{D}_{\text{ood}}$ of out-of-distribution examples \cite{degroot83evaluation, daxberger22ood}. Such examples may correspond, as discussed, to examples recorded in different lighting conditions with a neuromorphic camera. 
\subsection{Offline Learning}

\subsubsection{Two-moons dataset} 

We start by considering the two-moons dataset. For this experiment, the SNN comprises two fully connected layers with $256$ neurons each. Bayesian learning is implemented with different values of the temperature parameter $\rho$ in the free energy \eqref{eq:irm}. In Fig.~\ref{fig:twomoons_exp}, triangles indicate training points for a class ``$0$'', while circles indicate training points for a class ``$1$''. The color intensity represents the predictive probabilities \eqref{eq:pred-conf-freq} for frequentist learning and \eqref{eq:pred-conf-bayes} for Bayesian learning: the more intense the color, the higher the prediction confidence determined by the model. Bayesian learning is observed to provide better calibrated predictions, that are more uncertain outside the input regions covered by training data points. 

\begin{table}[t]
\caption{Final average test accuracy and ECE on the split-MNIST dataset (real-valued synapses).}
\label{tab:comparison-continual}
\begin{center}
\begin{small}
\begin{sc}
\begin{tabular}{ccr}
\toprule
 Model & Accuracy & ECE \\
 \midrule
TACOS \cite{soures21tacos} (Full Precision) &  $83.45 \pm 0.55$\% & N/A \\
  Frequentist  \cite{kirkpatrick17ewc}  & $77.19 \pm 0.65$\% & $0.39 \pm 0.01$ \\ 
  Bayesian Committee Machine & $\bm{85.44 \pm 0.16}$\% & $\bm{0.36 \pm 0.01}$ \\ 
  Bayesian Ensemble Decision  & $85.03 \pm 0.54$\% & $\bm{0.36 \pm 0.01}$ \\
\bottomrule
\end{tabular}
\end{sc}
\end{small}
\end{center}
\vspace{-0.5cm}
\end{table}

\begin{figure}[t]
\centering
\includegraphics[width=\linewidth]{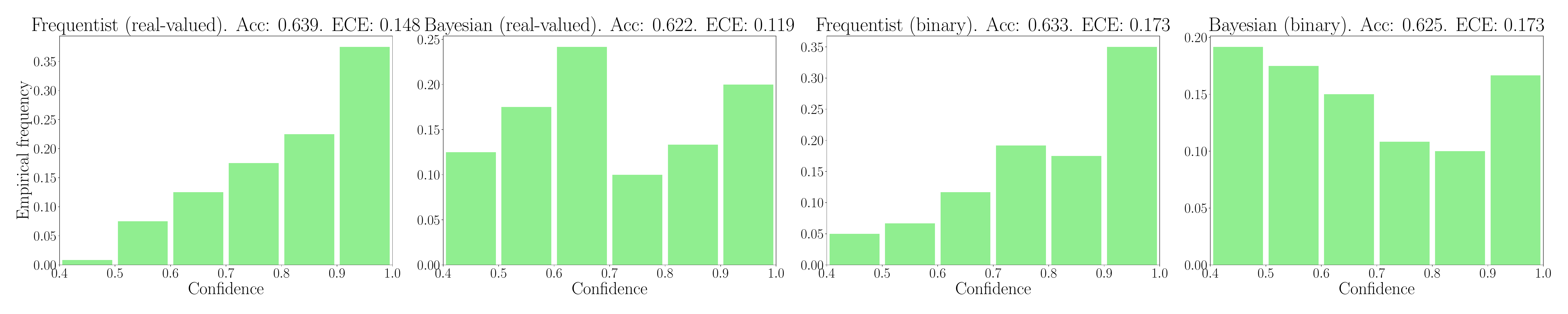}
\caption{Out-of-distribution empirical confidence histograms for SNNs with real-valued and binary synapses on the DVS-Gestures dataset. }
\label{fig:bayes-ood}
\end{figure}

For both real-valued and binary synapses, the temperature parameter $\rho$ has an important role to play in preventing overfitting and underfitting of the training data, while also enabling uncertainty quantification. When the parameter $\rho$ is too large, the model cannot fit the data correctly, resulting in inaccurate predictions; while when $\rho$ is too small, the training data is fit more tightly, leading to a poor representation of the prediction uncertainty outside the training set. A well-chosen value of $\rho$ strikes the best trade-off between faithfully fitting the training data and allowing for uncertainty quantification. Frequentist algorithms, obtained in the limit when $\rho \rightarrow 0$, yield the most over-confident estimates.

\subsubsection{DVS-Gestures}

We now turn to the DVS-Gestures dataset, for which we plot the performance for real-valued and binary-valued SNNs, in terms of accuracy, reliability diagrams \cite{degroot83evaluation}, and ECE \cite{guo2017hierarchical} in  Fig.~\ref{fig:bayes-id-real} and Fig.~\ref{fig:bayes-id-bin}. In all cases, the SNNs have two fully connected layers comprising $4096$ neurons each, and they are trained for $200$ epochs. The architecture was chosen to highlight the benefits of Bayesian learning over frequentist learning  in regimes characterized by epistemic uncertainty, and it was not optimized for maximal accuracy. The figures confirm that Bayesian SNNs generally produce better calibrated outcomes. In fact, reliability diagrams (top rows) demonstrate that frequentist learning algorithms produce overconfident decisions, while Bayesian learning outputs decisions whose confidence levels match well the test accuracies. This improvement is reflected, for models with real-valued synapses (with fixed precision), in a lower ECE of $0.064$, as compared to $0.088$ for frequentist SNNs; while, for binary SNNs, the reduction in ECE is from $0.085$ for frequentist learning, to $0.069$ for Bayesian learning. This benefit may come at the cost of a slight decrease in terms of accuracy, which is only observed here for binary synapses. The bottom parts of Fig.~\ref{fig:bayes-id-real} and Fig.~\ref{fig:bayes-id-bin} also show that frequentist learning tends to output high-confidence decisions with a larger probability. 

\begin{figure}[t]
\centering
\includegraphics[width=1.\textwidth]{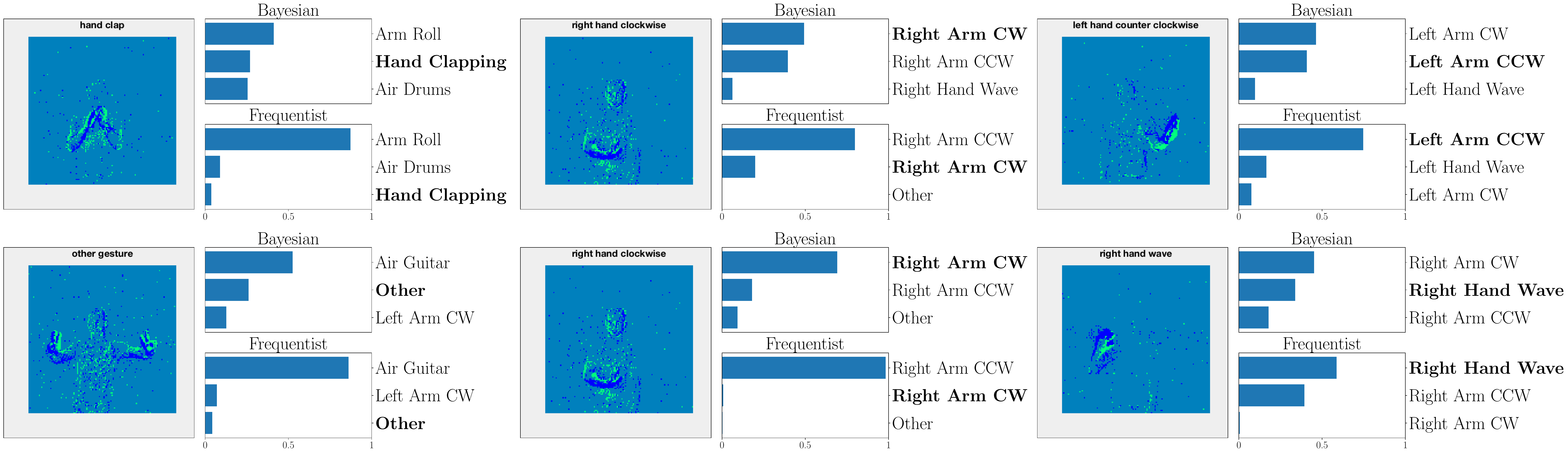} 
\caption{Top three classes predicted by both Bayesian and frequentist models on selected examples. Top: real-valued synapses. Bottom: binary synapses. The correct class is indicated in bold font.}
\label{fig:visualisation}
\end{figure}

We now turn to evaluate the performance in terms of robustness to out-of-distribution data. As explained in Sec.~\ref{sec:perf-meas}, to this end, we evaluate the histogram of the confidence levels produced by frequentist and Bayesian learning, as shown in Fig.~\ref{fig:bayes-ood}. From the figure, it is remarked that Bayesian learning correctly provides low confidence levels on out-of-distribution data, while frequentist learning outputs decisions with confidence levels similar to the case of in-distribution data, which are shown in Fig.~\ref{fig:bayes-id-real} and Fig.~\ref{fig:bayes-id-bin}. 

This point is further illustrated  in Fig.~\ref{fig:visualisation} by showing the three largest probabilities assigned by the different models on selected examples, considering real-valued synapses in the top row and  binary synapses in the bottom row. In the left column, we observe that, when both models predict the wrong class, Bayesian SNNs tend to do so with a lower level of certainty, and typically rank the correct class higher than their frequentist counterparts. Specifically, in the examples shown, Bayesian models with both real-valued and binary synapses rank the correct class second, while the frequentist models rank it third. Furthermore, as seen in the middle column, in a number of cases, the Bayesian models manage to predict the correct class, while the frequentist models predict a wrong class with high certainty. Finally, in the right column, we show that even when frequentist models predict the correct class and Bayesian models fail to do so, they still assign lower probabilities (i.e., $<50\%$) to the predicted class.

\begin{figure}[t]
\centering
\includegraphics[width=1.\textwidth]{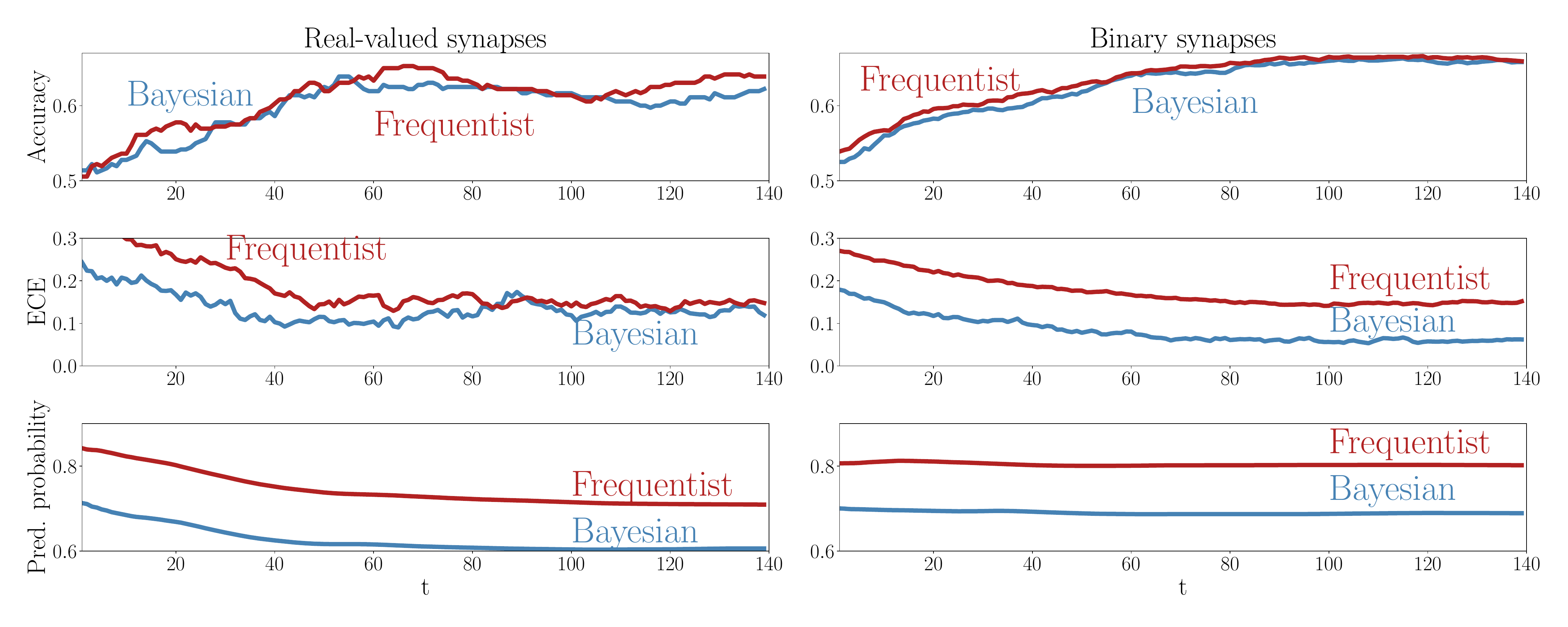} 
\caption{Evolution of the accuracy (top), ECE (middle), and predictive probabilities (bottom) during the presentation of out-of-distribution test examples for the DVS-Gestures dataset. The horizontal axis represents the time instants $t$ within the presentation of each test example. Left: Real-valued synapses. Right: Binary synapses.}
\label{fig:online}
\end{figure}

A key advantage of SNNs is the possibility to obtain intermediate decisions during the observation of the $T$ samples of a test example. To elaborate on this aspect, Fig.~\ref{fig:online} reports the evolution of the mean test accuracy, ECE, and predictive probabilities \eqref{eq:pred-conf-bayes}-\eqref{eq:pred-conf-freq} for all examples in the out-of-distribution dataset as a function of the discrete time-steps $t = 1, 2, \dots, T$. Although both Bayesian and frequentist methods show similar improvements in accuracy over time, frequentist algorithms remain  poorly calibrated, even after the observation of many time samples. The bottom plots show that frequentist learning tends to be more confident in its decisions, especially when a few samples $t$ have been observed. On the contrary, Bayesian algorithms offer better calibration and confidence estimates, even when only part of the input signal $\bmx$ has been observed. 

\subsection{Continual Learning}

We now turn to continual learning benchmarks. Starting with the rate encoded MNIST dataset, we use coresets representing $7.5$\% of randomly selected training examples for each class. To establish a fair comparison with the protocol adopted in reference~\cite{soures21tacos}, we train SNNs comprising a single layer with $400$ neurons for one epoch on each subtask. This choice was found to be advantageous for Bayesian techniques -- a result that may be related to the known asymptotic behavior of Bayesian neural networks as non-parametric models \cite{neal96bayesian}. In Table~\ref{tab:comparison-continual}, we show the average accuracy over all tasks at the end of training on the last task, as well as the average ECE at that point for real-valued synapses, enabling a comparison with \cite{soures21tacos}. Bayesian continual learning is seen to achieve the best accuracy and calibration across all the methods studied here, including the solution introduced in \cite{soures21tacos}. The latter incurs a $2.5\times$ memory overhead as compared to standard frequentist methods. Considering that we performed training using the $8$-bit precision imposed by the neuromorphic chip Loihi, our solution outperforms the state-of-the-art with a $5\times$ memory consumption improvement. This saving can be leveraged, e.g., to store several samples of the weights for a committee machine implementation.

\begin{figure}[t]
\centering
\includegraphics[width=1.\textwidth]{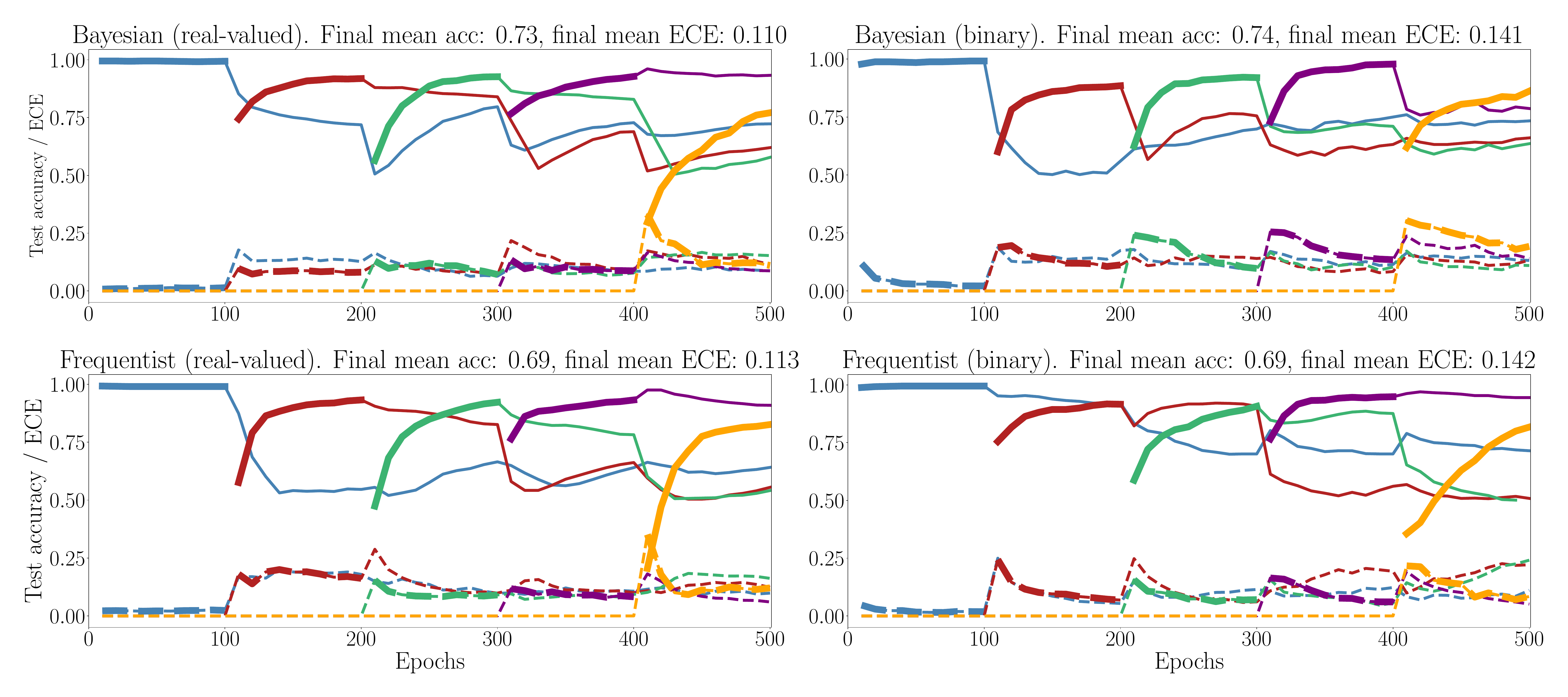} 
\caption{Evolution of the average test accuracies and ECE on all tasks of the split-MNIST-DVS across training epochs, with Gaussian and Bernoulli variational posteriors, and frequentist schemes for both real-valued and binary synapses. Continuous lines: test accuracy, dotted lines: ECE, bold: current task. Blue: $\{0, 1\}$; Red: $\{2, 3\}$; Green: $\{4, 5\}$; Purple:$\{6, 7\}$; Yellow: $\{8, 9\}$. }
\label{fig:continual-exp}
\end{figure}

Next, for the MNIST-DVS dataset \cite{serrano2015poker}, we use coresets representing $10$\% of randomly selected training examples for each class, and implement multilayer SNNs with $2048 - 4096 - 4096 - 2048 - 1024$ neurons per layer, that we train on each subtask for $100$ epochs. This task requires a larger architecture and longer training time to allow for the processing of the richer spatio-temporal information recorded by neuromorphic cameras, as compared to the spatial information from static image datasets, such as MNIST, encoded into spikes via rate encoding \cite{snnreview}.

\begin{figure}[t]
\centering
\includegraphics[width=1.\textwidth]{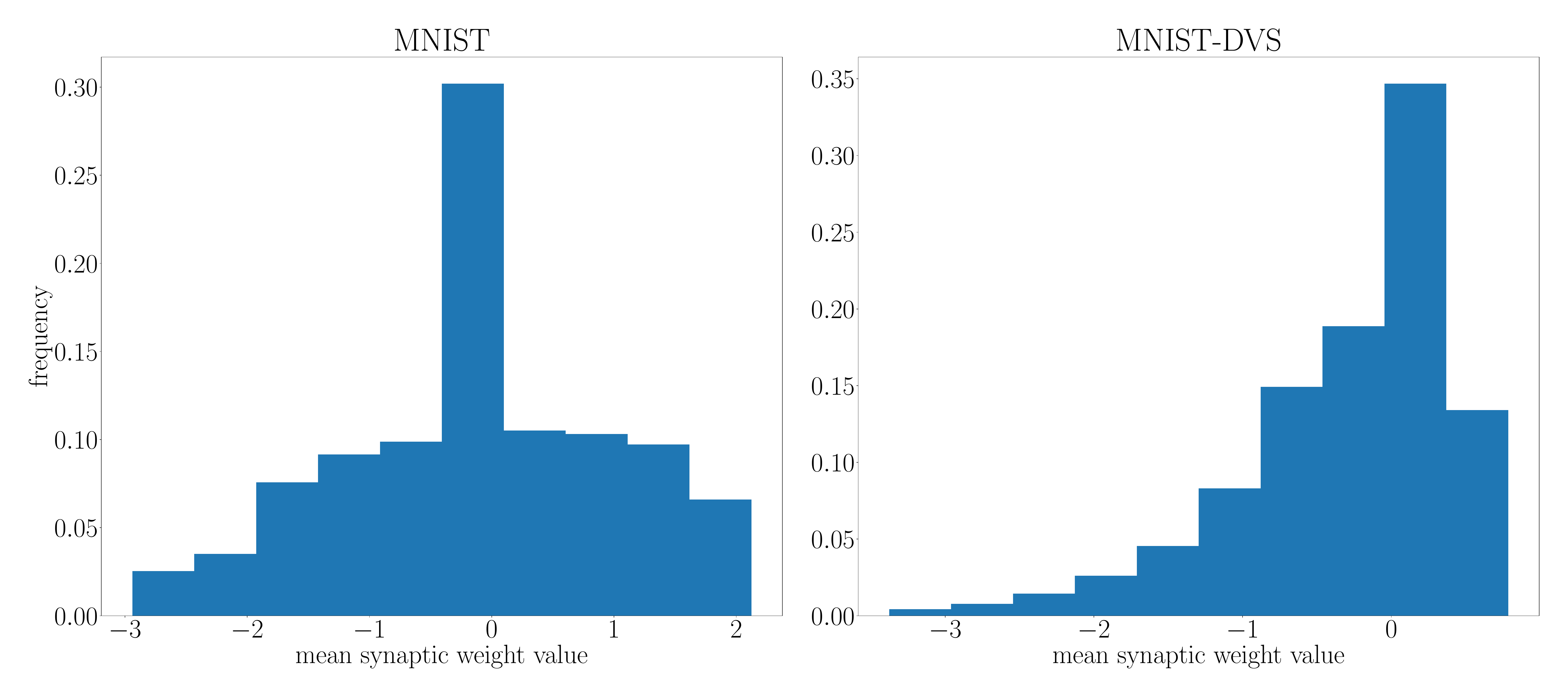} 
\caption{Distribution of the mean parameter $\bm{m}$ at the end of training on the MNIST and MNIST-DVS datasets. }
\label{fig:weights_distrib}
\end{figure}

We highlight the requirement for a larger architecture on the MNIST-DVS dataset in Fig.~\ref{fig:weights_distrib} by comparing the distribution of the mean parameter $\bm{m}$ at the end of training on the MNIST and MNIST-DVS datasets. For the larger network trained on the MNIST-DVS dataset, $83.5\%$ of the mean parameters are non-zero, a larger proportion than that of the network trained on the MNIST dataset, for which only $80.1\%$ of the mean weight parameters are non-zero. This demonstrates that the larger number of weights used for this task is important for the network to perform well.

In Fig.~\ref{fig:continual-exp}, we show the evolution of the test accuracy and ECE on all tasks, represented with lines of different colors, during training. The performance on the current task is shown as a thicker line. We consider frequentist and Bayesian learning, with both real-valued and binary synapses. With Bayesian learning, the test accuracy on previous tasks does not decrease excessively when learning a new task, which shows the capacity of the technique to tackle catastrophic forgetting. Also, the ECE across all tasks is seen to remain more stable for Bayesian learning as compared to the frequentist benchmarks. For both real-valued and binary synapses, the final average accuracy and ECE across all tasks show the superiority of Bayesian over frequentist learning. 

\begin{figure}[t]
\centering
\includegraphics[width=1.\textwidth]{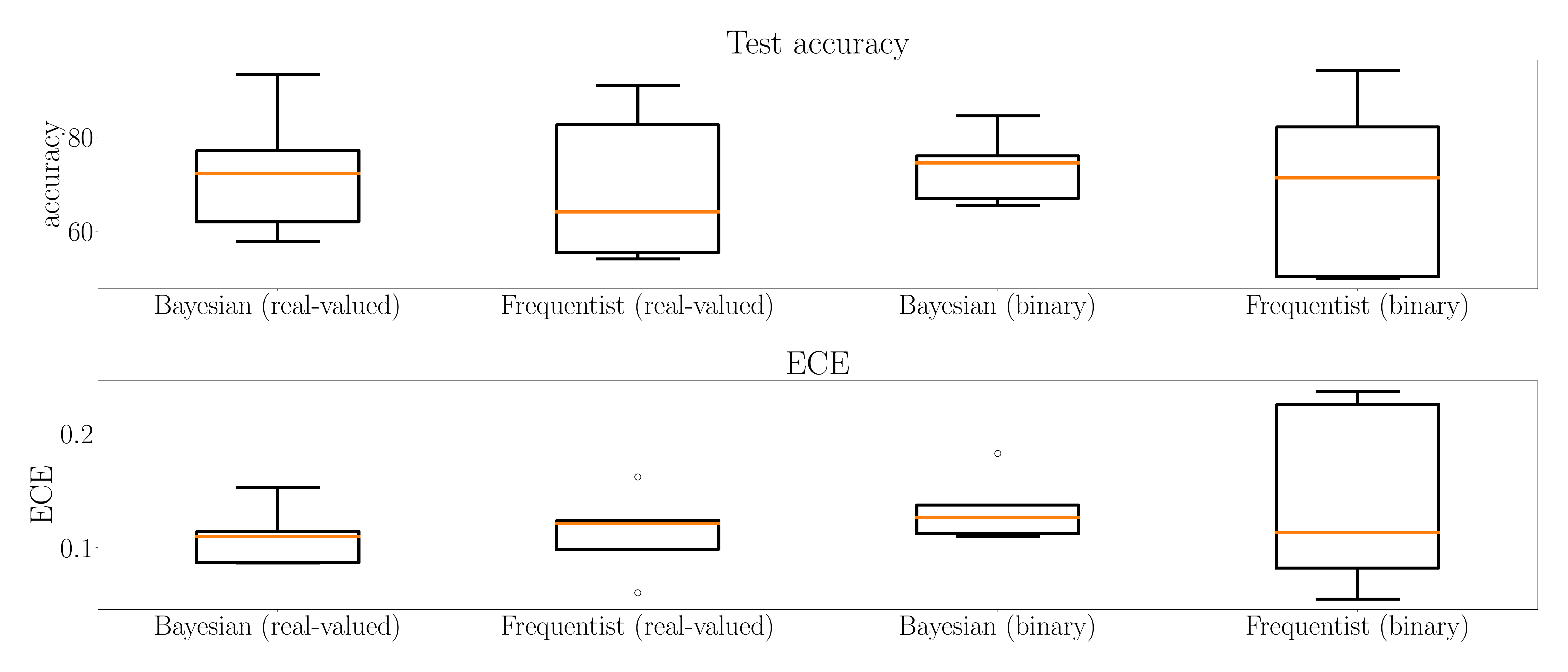} 
\caption{Box plots for final test accuracy (top) and ECE (bottom) across all tasks for Bayesian and frequentist learning with real-valued and binary synapses after the $500$th epoch of training on the MNIST-DVS dataset (see Fig.~\ref{fig:continual-exp}). The horizontal bar represents the median value across the tasks, while the box extends from the first to the third quartile. The whiskers extend from the box by $1.5$ times the inter-quartile range. Circles represent outliers.}
\label{fig:continual-final}
\end{figure}

This point is further elaborated in Fig.~\ref{fig:continual-final}, which shows test accuracy and ECE on all tasks at the final epoch -- the $500$th -- in Fig.~\ref{fig:continual-final}.  Bayesian learning can be seen to offer a better test accuracy and ECE on average across tasks, as well as a lower dispersion among tasks. 

\section{Conclusion}
In this work, we have introduced a Bayesian learning framework for SNNs with both real-valued and binary-valued synapses. Bayesian learning is particularly well suited for applications characterized by limited data -- a situation that is likely to be encountered in use cases of neuromorphic computing such as edge intelligence. We have demonstrated the benefits of Bayesian learning in terms of calibration metrics that gauge the effectiveness of uncertainty quantification over a variety of offline and continual learning. We have also argued that the proposed rules exhibit mechanisms resembling those that enable lifelong learning in biological brains from a theoretically motivated information risk minimization framework. While this work focused on variational inference Bayesian learning methods, future research may explore Monte-Carlo based solutions. Finally, we recall the importance of investigating solutions for hardware design, adopting either ensemble predictors or committees of machines. As an example, consider ensemble predictions based on binary synapses. An implementation based on digital hardware would need to store the real-valued parameters of the parameter vector distribution, and to sample from the distribution using auxiliary circuitry, which incurs energy and memory overheads. Alternatively, one could leverage the inherent stochasticity of analog hardware for sampling \cite{dalgaty21memristor}, a line of research that we reserve for future work.

\section*{Author Contributions}
OS first proposed to train SNNs via Bayesian learning, HJ derived the rule for binary synapses, and NS extended it to real-valued synapses. NS designed and implemented the experiments. NS, HJ, and OS wrote the text.

\section*{Funding}
This study received funding from Intel Labs through the Intel Neuromorphic Research Community (INRC). The work of O. Simeone has also been supported by the European Research Council (ERC) under the European Union's Horizon 2020 Research and Innovation Programme (Grant Agreement No. 725731). The funders were not involved in the study design, collection, analysis, interpretation of data, the writing of this article or the decision to submit it for publication. All authors declare no other competing interests.

\section*{Data Availability Statement}
The datasets analyzed for this study are freely available and can be found at references \cite{twomoonsdataset, lecun89handwritten, serrano2015poker}.

\bibliographystyle{IEEEbib}
\bibliography{ref}

\newpage
\section{Supplementary Material}
Hyperparameters chosen for the experiments are summarized in Table~\ref{tab:parameters}.

\begin{landscape}
 \label{tab:parameters}
\begin{tabular}{ |p{3cm}||p{2.5cm}|p{2.5cm}|p{2cm}|p{2cm}|p{2cm}|p{2.5cm}|p{1.5cm}|p{1.5cm}|}
\hline
 \multicolumn{9}{|c|}{Hyperparameters} \\
 \hline
 Dataset & Algorithm & Synapses &  Number of epochs & Batch size & $\eta$ & $\rho$ & $\alpha$ & $\tau$\\
 \hline
  Two-Moons & Bayesian & Real-valued & $100$ & $64$ & $0.01$ & 1e-12 -- 1e-1 & N/A & N/A \\
  & Frequentist & Real-valued & $100$ & $64$ & $0.001$ & N/A & N/A & N/A \\
  & Bayesian & Binary & $100$ & $64$ & $0.1$ & 1e-12 -- 1e-1 & N/A & $1.$ \\
  & Frequentist & Binary & $100$ & $64$ & $0.001$ & N/A & N/A & N/A \\
  \hline
 
  DVSGestures & Bayesian & Real-valued & $200$ & $64$ & $0.02$ & $0.001$ & N/A & N/A \\
  & Frequentist & Real-valued & $200$ & $64$ & $0.001$ & N/A & N/A & N/A \\
  & Bayesian & Binary & $200$ & $64$ & $0.004$ & $0.001$ & N/A & $1.$ \\
  & Frequentist & Binary & $200$ & $64$ & $0.00002$ & N/A & N/A & N/A \\
  \hline
  
  MNIST & Bayesian & Real-valued & $1$ & $64$ & $0.5$ & $0.000075$ & N/A & N/A \\
  & Frequentist & Real-valued & $1$ & $64$ & $0.0003$ & N/A & $1.$ & N/A \\
  \hline
  
  MNIST-DVS & Bayesian & Real-valued & $100$ & $64$ & $0.5$ & $0.00001$ & N/A & N/A \\
  & Frequentist & Real-valued & $100$ & $64$ & $0.003$ & N/A & $1.$ & N/A \\
  & Bayesian & Binary & $100$ & $64$ & $0.1$ & $0.000004$ & N/A & $1.$ \\
  & Frequentist & Binary & $100$ & $64$ & $0.0000006$ & N/A & $1.$ & N/A \\
 \hline
\end{tabular}
\end{landscape}

\end{document}